\documentclass[10pt,twocolumn,letterpaper]{article}

\usepackage[pagenumbers]{iccv} 

\usepackage[dvipsnames]{xcolor}

\definecolor{iccvblue}{rgb}{0.21,0.49,0.74}
\usepackage[pagebackref,breaklinks,colorlinks,allcolors=iccvblue]{hyperref}
\usepackage{enumitem}
\usepackage{algorithm}
\usepackage{algpseudocode}
\usepackage{amsmath}
\usepackage{graphicx}
\usepackage{dsfont}
\usepackage{multirow}

\usepackage{tcolorbox}
\tcbuselibrary{most}
\newtcolorbox{userprompt}[1]{
    enhanced,
    drop shadow=black!5!white,
    left=4mm,
    right=4mm,
    top=3mm,
    bottom=3mm,
    boxsep=0mm,
    rounded corners,
    title=#1,
    fontupper=\linespread{1.1}\scriptsize\fontfamily{lmr}\selectfont,
    breakable
}

\def\paperID{4755} 
\def\confName{ICCV}
\def\confYear{2025}

\title{Discovering Divergent Representations between Text-to-Image Models}

\author{
Lisa Dunlap$^{1}$\footnotemark[1] \quad
Joseph E. Gonzalez$^{1}$ \quad
Trevor Darrell$^{1}$ \quad
Fabian Caba Heilbron$^{2}$ \quad \\
Josef Sivic$^{2,3}$ \quad
Bryan Russell$^{2}$ \\
$^1$University of California, Berkeley \quad
$^2$Adobe Research \quad
$^3$CIIRC CTU 
}
\newcommand{\attribute}{diverging visual attribute\xspace}
\newcommand{\dprompt}{diverging prompt\xspace}
\newcommand{\promptdescription}{diverging prompt description\xspace}

\newcommand{\dataset}{ID$^2$\xspace}
\newcommand{\method}{\textsc{CompCon}\xspace}  
\newcommand{\imageseta}[1]{\mathcal{I}_1^{(#1)}}
\newcommand{\imagesetb}[1]{\mathcal{I}_2^{(#1)}}

\newcommand{\code}[1]{\texttt{#1}}

\begin{document}



\twocolumn[{%
\renewcommand\twocolumn[1][]{#1}%
\maketitle
\begin{center}
    \centering
    \captionsetup{type=figure}    \includegraphics[width=\textwidth]{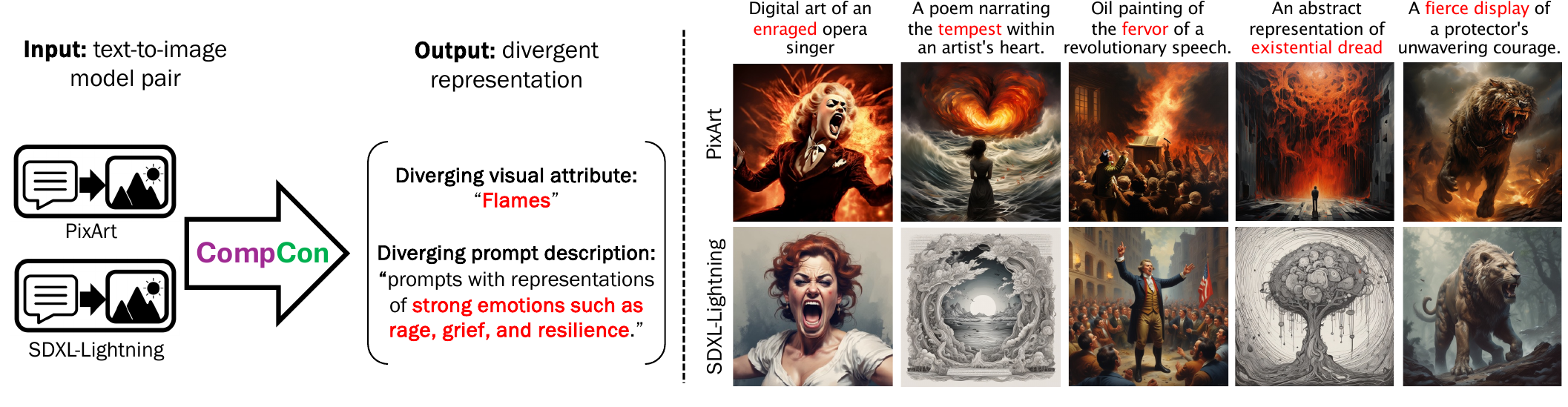}
    \vspace{-2.5em}
    \captionof{figure}{\textbf{Discovering divergent representations with \method}. {\em Left:} \method{} takes as input a pair of text-to-image models and outputs a \promptdescription{} to produce a \attribute{} appearing in one model but not the other. {\em Right:} We show the discovered \attribute{} `flames' appearing in PixArt but not SDXL-Lightning over different \dprompt{}s.}
    \label{fig:teaser}
\end{center}%
}]
\footnotetext[1]{Work done on internship at Adobe Research. Accepted to ICCV 2025.}

\begin{abstract}
In this paper, we investigate when and how visual representations learned by two different generative models {\bf diverge}. 
Given two text-to-image models, our goal is to discover visual attributes that appear in images generated by one model but not the other, along with the types of prompts that trigger these attribute differences. For example, 
`flames' might appear in one model’s outputs when given prompts expressing strong emotions, while the other model does not produce this attribute given the same prompts.
We introduce CompCon (Comparing Concepts), an evolutionary search algorithm that discovers visual attributes more prevalent in one model's output than the other, and uncovers the prompt concepts linked to these visual differences. 
To evaluate CompCon's ability to find diverging representations, we create an automated data generation pipeline to produce ID$^2$, a dataset of 60 input-dependent differences, and compare our approach to several LLM- and VLM-powered baselines. 
Finally, we use CompCon to compare popular text-to-image models, finding divergent representations such as how PixArt depicts prompts mentioning loneliness with wet streets and Stable Diffusion 3.5 depicts African American people in media professions. 
Code: \url{https://github.com/adobe-research/CompCon}.

\end{abstract}

\section{Introduction}


Generative models develop unique representations of semantic concepts -- for instance, happy scenes contain warm colors, or dogs are found outside. While many of these representations are shared across models, understanding when representations \textbf{diverge} can reveal stylistic differences between models. 
In this work, we explore how to uncover such divergent representations by identifying input-dependent differences between two text-to-image models. Specifically, we aim to discover pairs of semantic concepts and visual attributes where prompts containing a semantic concept cause one model to generate images displaying the corresponding visual attribute, while the other does not. For example, prompts that mention strong emotions result in images with flames in one model but not the other (Fig.~\ref{fig:teaser}).

Discovering divergent representations is beneficial for both model developers and users. For developers, it can help decide which model to deploy to production based on any problematic discovered divergent representations, and for evaluating against competitor models. For users, it can help in selecting the model that best aligns with their own interpretations and needs. Manually performing this task is labor-intensive, as it requires sifting through hundreds or thousands of images to find visual attribute differences; once identified, additional effort is needed to determine the types of input prompts that trigger these differences.

When comparing text-to-image models, the evaluation typically focuses on metrics such as image quality and prompt adherence~\cite{heusel2017gans,salimans2016improved,hessel2021clipscore,park2024benchmark,Grimal_2024_WACV,hu2023tifa,lee2023holistic}. While these metrics indicate \emph{how well} models perform, they often overlook \emph{what} the models actually learn. For example, what defines ‘cute’ versus ‘ugly’? What characteristics make something appear ‘futuristic’? What does `emotion' look like? As we will show, models trained on different data, using different encoders or training procedures, can learn distinct interpretations of the same concept. For instance, one model may associate `ancient' with the Paleolithic Era while the other associates it with the Roman Empire. 


To address the task of discovering divergent representations, we make the following contributions. First, we introduce \method{} (Comparing Concepts), an evolutionary search algorithm designed to uncover input-dependent differences in model representations. \method{} first discovers pairwise differences in model outputs, generates a description of the prompts that cause this difference, and iteratively refines this description by analyzing existing prompts and generating new ones likely to highlight these differences. As shown in Figure~\ref{fig:teaser}, \method{} can generate prompts that result in model behaviors like putting flames behind opera singers for one model but not the other. 

Second, we create \dataset{} (Input-Dependent Differences), a dataset of 60 semantic-visual representations to evaluate the efficacy of our method. Using this dataset, we compare \method{} to LLM, TF-IDF, and VisDiff~\cite{VisDiff} baselines. As our third contribution, we apply \method{} to compare the PixArt and SD-Lightning text-to-image models, finding, for example, that prompts mentioning anger result in depictions of `flames' in PixArt, prompts mentioning sadness and solitude result in `wet streets' in PixArt, and abstract prompts with cosmic motifs result in `mandala circular designs' in SD-Lightning. We also uncover bias, such as PixArt generating older men for prompts mentioning traditional professions. These findings demonstrate how \method{} can systematically reveal subtle differences between generative models, helping developers and users better understand and leverage these models’ unique behaviors.




\section{Related Work}

\noindent\textbf{Evaluating Text-to-Image Models.} The evaluation of text-to-image models has advanced significantly in recent years. Traditional quality measures such as FID~\cite{heusel2017gans}, Inception core~\cite{salimans2016improved}, CLIP score~\cite{hessel2021clipscore}, and CLIP-R score~\cite{park2024benchmark} have been complemented by newer metrics like TAIM~\cite{Grimal_2024_WACV} and TIFA~\cite{hu2023tifa}, which leverage vision-and-language models to assess prompt adherence. Holistic benchmarks such as HEIM and others ~\cite{lee2023holistic, cho2022dall,saharia2022photorealistic, yu2022scaling} aggregate multiple axes of evaluation, including quality, prompt adherence, style, and efficiency. While these approaches excel in measuring overall model performance, they focus on objective qualities and often overlook fine-grained, subjective differences between models. 
Our work complements these efforts by targeting input-dependent differences, particularly in semantic interpretations and stylistic variations, which are crucial for understanding model-specific behaviors.

\noindent\textbf{Interpreting Diffusion Models.}
Several works have explored ways of interpreting the internal representations of generative vision-language models. \citet{ gandelsman2023interpreting, netdissect,gandelsman2024neurons, Dravid_2023_ICCV} explore how to describe the function of certain neurons and attention heads in natural language,, and \citet{tong2023massproducing} discovers how image and text representations differ in latent space to better understand CLIP failures. We see our work as a data-driven approach to attain similar insights into model behavior. 

\noindent\textbf{Discovering Bias in Diffusion Models.} Uncovering and mitigating biases in text-to-image models has been well explored, with many works focus on finding and mitigating a predefined set of biases related to gender, race, and geography ~\cite{bianchi2023easily,wang2023t2iatmeasuringvalencestereotypical, wang2020towards,hamidieh2023identifying,ghosh2023person,friedrich2023fair,cho2023dalleval}. Recently, a line of works have emerged that aim to automatically discover biases from the data, rather than using a predefined list. Many of these approaches \cite{D'Inca_2024_CVPR,dincà2024gradbiasunveilingwordinfluence} discover bias by prompting a large language model (LLM) to propose potential biases from image captions, generating prompts that may indicate a bias in models (\eg, ``a doctor") and using a VLM to check if this bias exists. \citet{liu2024organizing} builds on this by clustering generated images based on concepts like gender, while \citet{chinchure2023tibet} extends bias discovery to counterfactual examples, eliminating the need for a large caption pool.
In contrast, our work focuses on model \emph{comparison}, rather than single-model auditing, which better reflects many real-world evaluations where success is measured by improvement relative to other models. Additionally, while previous methods \cite{D'Inca_2024_CVPR,dincà2024gradbiasunveilingwordinfluence} identify biases from input captions, CompCon analyzes generated images directly using VLMs. This enables discovery of subtler, more nuanced differences in visual representation, including social biases and stylistic or conceptual divergences between models. Further discussion is in Sections~\ref{sec:pixart_sd_lightning} and~\ref{sec:compare_to_openbias_supp}.




\noindent\textbf{Describing Differences in Image Sets.} Several works have aimed to describe differences in sets of images using natural language. 
For example, VisDiff~\cite{VisDiff} generates visual attributes distinguishing two sets of images by analyzing captions and refining them using cross-modal embeddings. Similarly, \citet{chiquier24llmmutate} train interpretable CLIP classifiers that evolve based on LLM-generated attributes. While these approaches focus on dataset-level differences, they do not address input-dependent variation. We adapt and extend VisDiff’s methodology for pairwise comparison, focusing on prompt-specific divergences and refining attribute discovery to capture subtler differences. Additionally, our method introduces a novel iterative search for prompt descriptions that cause these differences.

\section{Divergent Representation Discovery}
\label{sec:method}

\begin{figure*}
    \centering
    \includegraphics[width=\linewidth]{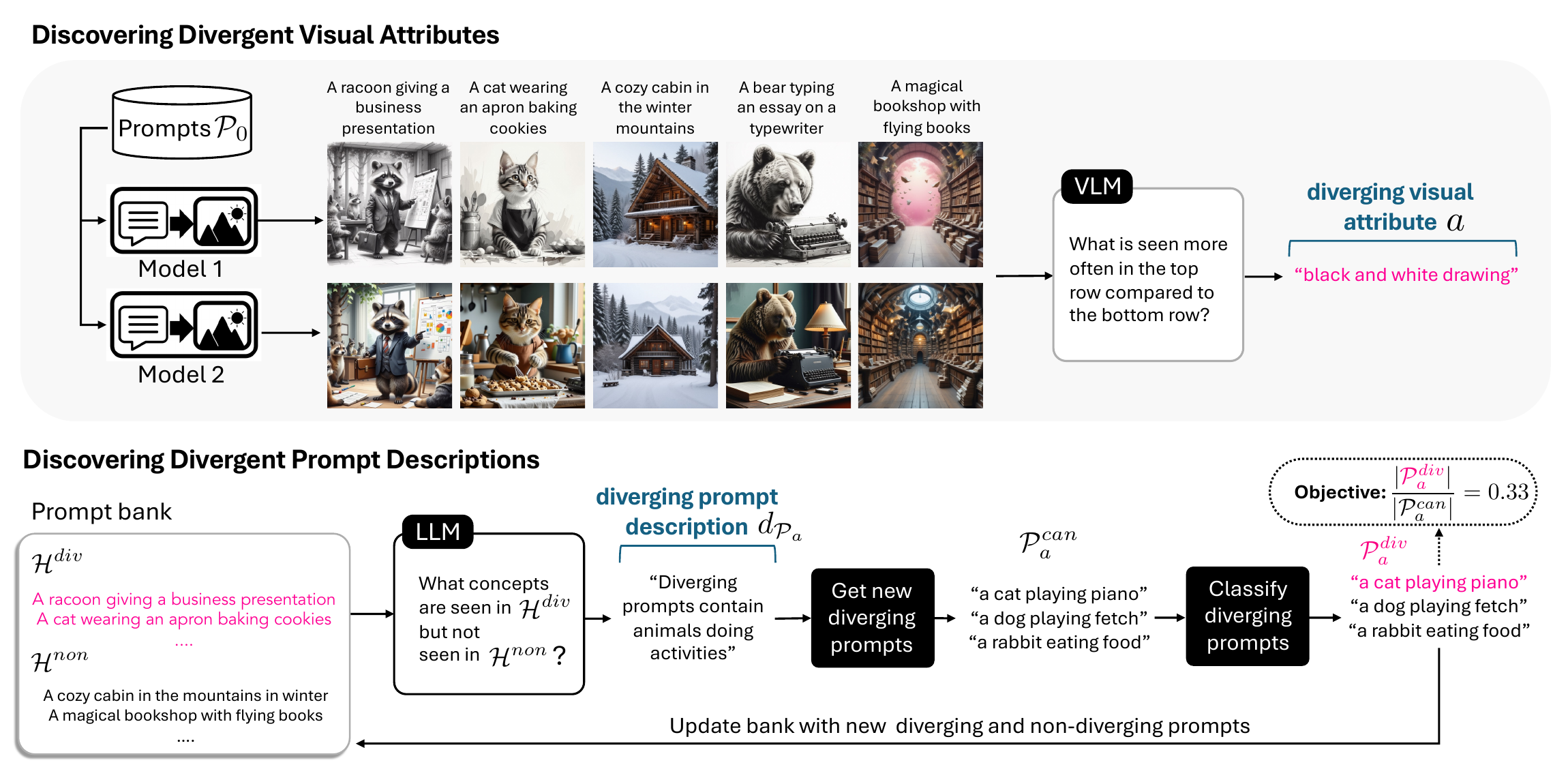}
    \vspace{-2em}
    \caption{\textbf{\method{} overview.} We illustrate our approach for discovering \attribute{}s (top) and \promptdescription{}s (bottom). Given two text-to-image models and a set of prompts, we use a VLM to identify visual differences. For each diverging attribute, we iteratively refine  diverging prompt descriptions by generating candidate prompts $\mathcal{P}_a^{can}$ from the description, classifying them as diverging ($\mathcal{H}^{div}$) or non-diverging ($\mathcal{H}^{non})$. The objective is to maximize the proportion of generated prompts classified as diverging.}
    \label{fig:method_overview}
    \vspace{-1em}
\end{figure*}

Let $P$ be a set of text prompts, and $\mathcal{I}_1^{(P)}$ and $\mathcal{I}_2^{(P)}$ be sets of images generated by two text-to-image models given the input prompts $P$. 
We call the natural language descriptions of the prompt set $P$ and any visual differences between the two generated image sets $\mathcal{I}_1^{(P)}$ and $\mathcal{I}_2^{(P)}$ a {\em divergent representation} (see Figure~\ref{fig:teaser}).
These divergent representations take the form of a pair of natural language descriptions $(a, d_{P_a})$, where $a$ is a  description of a visual attribute seen more often in one model than the other (\eg, ``flames''), and $d_{P_a}$ is a description of the concepts present in text prompts $P_a$ eliciting this difference (\eg, ``strong emotions''). 
We refer to $a$ as a \emph{\attribute{}}, $d_{P_a}$ as a \emph{\promptdescription{}}, and $P_a$ as \emph{\dprompt{}s}. 

Given a pair of text-to-image models $\Theta = (\theta_1, \theta_2)$, we aim to discover differences in the images generated by the two models as well as a description of the types of input text prompts that trigger these differences.
Let $\mathcal{P}$ be the set of all possible text prompts and  $\mathcal{A}$ be the set of all possible \attribute{}s. Our goal is to find the mapping $\mathcal{F}_\Theta$ from text prompts in $\mathcal{P}$ to \attribute{}s in $\mathcal{A}$ given the model pair $\Theta$,
\begin{equation}
    \mathcal{F}_\Theta : \mathcal{P} \mapsto \mathcal{A}.
\end{equation}
Note that this mapping is not a bijection as multiple text prompts may map to a single \attribute{}. Moreover, multiple \attribute{}s may be depicted for a given set of \dprompt{}s.

Our approach for computing the mapping $\mathcal{F}_\Theta$ comprises two steps, illustrated in Figure~\ref{fig:method_overview}. First, given the text-to-image model pair $\Theta$ and a large set of initial prompts $\mathcal{P}_0 \subset \mathcal{P}$, we compute a set of \attribute{}s $\mathcal{A}_0 \subset \mathcal{A}$ (Section~\ref{sec:attribute}). Next, for each \attribute{} $a\in\mathcal{A}_0$, we optimize an objective to find the set of \dprompt{}s $\mathcal{P}_a \subset \mathcal{P}$ resulting in the \attribute{} $a$ (Section~\ref{sec:prompt}). We next describe each of these steps.

\subsection{Discovering Diverging Visual Attributes}
\label{sec:attribute}

Our goal is, given a text-to-image model pair $\Theta$ and a large set of initial text prompts $\mathcal{P}_0$, to compute a set of diverging visual attributes $\mathcal{A}_0$ over images generated given the prompts. This task is challenging as a system must identify any consistent visual differences between the two models' generated image sets. These differences are often subtle and difficult to spot over the large generated image collection. We address this challenge by prompting an off-the-shelf vision-language model (VLM) for this task.

 For the text-to-image model pair $\Theta$ and a text prompt $p$, let $\mathcal{G}(\Theta, p) = \left\{\mathcal{I}_1^{(p)}, \mathcal{I}_2^{(p)}\right\}$ denote the two sets of images generated by each model given prompt $p$. We first sample a batch of prompts $\mathcal{P}_{batch} \subset \mathcal{P}_0$ and, for each prompt $p\in \mathcal{P}_{batch}$, we construct a two-row image grid by tiling the images in $\mathcal{I}_1^{(p)}$ on the top row and $\mathcal{I}_2^{(p)}$ on the bottom row. Using this image grid, we instruct a VLM to find \attribute{}s appearing more in images of $\mathcal{I}_1^{(p)}$ compared to $\mathcal{I}_2^{(p)}$ (see Appendix for our instruction prompt).
 Our resulting \attribute{} list $\mathcal{A}_0$ is the aggregation of discovered attributes across $\mathcal{P}_{batch}$. 
 
 Next, we rank each \attribute{} $a\in\mathcal{A}_0$ by assigning a score indicating how well attribute $a$ can distinguish image sets $\mathcal{I}_1^{(\mathcal{P}_0)}$ and $\mathcal{I}_2^{(\mathcal{P}_0)}$. For each set of images generated by prompt $p\in \mathcal{P}_0$, we define a \emph{divergence score} $z(a, \mathcal{I}_1^{(p)}, \mathcal{I}_2^{(p)}) \to \{0, 1\}$ that indicates whether image set $\mathcal{I}_1^{(p)}$ contains attribute $a$ while  $\mathcal{I}_2^{(p)}$ does not. 
 
 Using cross-modal similarity, here CLIP~\cite{radford2021clip}, we compute the cosine similarity $s(\cdot)$ between \attribute{} $a$ and each image in sets $\mathcal{I}_1^{(p)}$ and $\mathcal{I}_2^{(p)}$. Using these similarities, we define the divergence score as the product of two indicated conditions,
\begin{align}
    z(a, \mathcal{I}_1^{(p)}, \mathcal{I}_2^{(p)}) =& 
     \mathds{1} \big[ s(a, \mathcal{I}_1^{(p)}) > t] \ \times \label{eqn:div_score} \\
     & \mathds{1} \big[ s(a, \mathcal{I}_1^{(p)}) - s(a, \mathcal{I}_2^{(p)}) > \delta \big] \notag
\end{align}
where $t$ and $\delta$ are hyperparameters that determine if $\mathcal{I}_1^{(p)}$ contains attribute $a$ and $\mathcal{I}_2^{(p)}$ does not contain $a$. 

Using this divergence score, we define the overall score for attribute $a$ as the mean divergence score over prompts in the initial prompt set: $\frac{1}{|\mathcal{P}_0|} \sum_{p\in\mathcal{P}_0} z(a, \mathcal{I}_1^{(p)}, \mathcal{I}_2^{(p)})$. 
A score of 1 means that all images generated by model $\theta_1$ contain attribute $a$ and none of the images generated by $\theta_2$ contain $a$. A score of 0 indicates that $a$ never appears more often in images generated by $\theta_1$. Note that we are not optimizing for scores close to 1, we are simply interested in any attribute $a$ that obtains a score sufficiently above zero. Finally, as many attributes  $\mathcal{A}_0$ are semantically equivalent (e.g. "flames" and "fire") we prompt an LLM to remove similar attributes. 

\begin{algorithm}[t]
\small
\renewcommand{\algorithmicrequire}{\textbf{Input:}}
\renewcommand{\algorithmicensure}{\textbf{Output:}}
\caption{Discovering \promptdescription{}s}
\begin{algorithmic}[1]
\Require Model pair $\Theta$, \attribute{} $a$, initial set of prompts $\mathcal{P}_0$, number of iterations $N$
\Ensure Diverging prompt description $d_{\mathcal{P}_a}^\star$
\vspace{0.5em}
\State Initialize: prompt bank $\mathcal{H} \gets \emptyset$, scores $\sigma \gets \code{emptyArray}(N)$, descriptions $d_{\mathcal{P}_a} \gets \code{emptyArray}(N)$
\State $\mathcal{P}_a \gets \code{classifyDiverging}(\Theta, a, \mathcal{P}_0)$
\State $\mathcal{H}^{div} \gets \mathcal{P}_a$
\State $\mathcal{H}_a^{non} \gets \mathcal{P} \setminus \mathcal{P}_a$
\vspace{0.5em}
\For{$i$ in $1,\dots,N$}
    \State $h_a^{div}, h_a^{non} \gets \code{sample}(\mathcal{H}^{div}, \mathcal{H}^{non})$
    \State $d_{\mathcal{P}_a}[i] \gets \code{describeDiverging}
    (\mathcal{P}_a^{div}, \mathcal{P}_a^{non}, \mathcal{H})$
    \State $\mathcal{P}_a^{can} \gets \code{getNewDiverging}(d_{\mathcal{P}_a}[i],a,\mathcal{H})$
    \State $\mathcal{P}_a^{div}, \mathcal{P}_a^{non} \gets \code{classifyDiverging}(\Theta, a, \mathcal{P}_a^{can})$
    \State $\sigma[i] \gets |\mathcal{P}_a^{div}| / |\mathcal{P}_a^{can}|$
    \State $\mathcal{H}^{non} = \mathcal{H}^{non} \cup \mathcal{P}_a^{non}$
    \State $\mathcal{H}^{div} = \mathcal{H}^{div} \cup \mathcal{P}_a^{div}$
\EndFor
\State $d_{\mathcal{P}_a}^\star \gets d_{\mathcal{P}_a}[\code{argmax}(\sigma)]$
\vspace{0.5em}
\State \Return $d_{\mathcal{P}_a}^\star$
\end{algorithmic}
\label{alg:method}
\end{algorithm}

\subsection{Discovering Diverging Prompt Descriptions}
\label{sec:prompt}

After we discover a set of \attribute{}s $\mathcal{A}_0$, for each attribute $a \in \mathcal{A}_0$ we aim to find a natural language description $d_{\mathcal{P}_a}$ of the \dprompt{}s that trigger this attribute. 
This task is challenging as the search space over all possible text prompts $\mathcal{P}$ is large, and we must not only find a set of prompts but a natural language description that completely covers this set.  

Let $\mathcal{L}(d_{\mathcal{P}_a})$ be the \dprompt{}s generated from description $d_{\mathcal{P}_a}$. Our objective is to maximize the expected divergence score over the generated prompts:

\begin{equation}
    \max_{d_{\mathcal{P}_a}} \mathbb{E}_{p\sim \mathcal{L}(d_{\mathcal{P}_a})}[z(a, \mathcal{I}_1^{(p)}, \mathcal{I}_2^{(p)})], 
    \label{eqn:prompt_objective}
\end{equation}
where $z$ is the divergence score defined in Equation~(\ref{eqn:div_score}).
That is, we want to maximize the number of prompts that have been generated by description $d_{\mathcal{P}_a}$ and confirmed to be diverging. We explore two options for our prompt generator $\mathcal{L}$. The first is to use an LLM to generate new prompts given $d_{\mathcal{P}_a}$, and the second is to retrieve prompts from a large prompt bank using $d_{\mathcal{P}_a}$. 

As the desired \promptdescription{} $d_{\mathcal{P}_a}$ is discrete (natural language) and Objective~(\ref{eqn:prompt_objective}) is not differentiable, we optimize the objective via evolutionary search. To achieve this goal, for $N$ evolutionary search iterations, we use an LLM to generate description $d_{\mathcal{P}_a}$ given a bank of diverging and non-diverging prompts $\mathcal{H}^{div}$ and $\mathcal{H}^{non}$, generate new prompts and images from $d_{\mathcal{P}_a}$, add these new prompts to $\mathcal{H}^{div}$ and $\mathcal{H}^{non}$, and evolve our description using these updated sets.

We maintain a bank $\mathcal{H}=\left(\mathcal{H}^{div}, \mathcal{H}^{non}\right)$ of diverging and non-diverging text prompts, where diverging prompts are given by $\mathcal{H}^{div} = \{p \;|\; z(a, \imageseta{p}, \imagesetb{p}) = 1\}$ and non-diverging prompts are given by $\mathcal{H}^{non} = \{p \;|\; z(a, \imageseta{p}, \imagesetb{p}) = 0\}$. We mutate the current description $d_{\mathcal{P}_a}$ by prompting an LLM (GPT-4o) to provide a description of what concepts are shared in $\mathcal{H}^{div}$ but not in $\mathcal{H}^{non}$ and score the mutation using Objective~(\ref{eqn:prompt_objective}). The mutation prompt is provided in the Appendix.

Algorithm~\ref{alg:method} provides pseudocode for our evolutionary search algorithm, with its key functions described below. 

\noindent
{\bf Mutation function} \code{describeDiverging}{\bf .}
Given the bank of diverging and non-diverging text prompts $\mathcal{H}$, we sample $B$ prompts from $\mathcal{H}^{div}$ and $\mathcal{H}^{non}$ and instruct an LLM to output a description $d_{\mathcal{P}_a}$ (the diverging prompt description) of what concepts are shared across diverging prompts which are not seen in non-diverging prompts. After the first iteration, prompts in $\mathcal{H}$ which were generated in the previous iteration are up-weighted when sampling.


\noindent
{\bf Mutation scoring.} 
To score the current mutation, we first define a function \code{getNewDiverging} that, given the \promptdescription{} $d_{\mathcal{P}_a}$, \attribute{} $a$, and prompt bank $\mathcal{H}$, provides candidate prompts $\mathcal{P}_a^{can}$ that are likely to be diverging and do not directly relate to attribute $a$. We explore two ways of obtaining $\mathcal{P}_a^{can}$: generation and retrieval. In the generation setting, we prompt an LLM (GPT-4o) to generate a diverse set of $k$ new prompts that align with the description $d_{\mathcal{P}_a}$, given random samples of prompts from $\mathcal{H}$ as a point of reference. The instruction prompt can be found in the Appendix. In the retrieval setting we use description $d_{P_a}$ to retrieve the top $k$ prompts from the prompt bank $\mathcal{H}$ excluding the $B$ sampled prompts used to generate the prompt description, having the highest text embedding similarity to $d_{P_a}$. We show results of both of these approaches in Section~\ref{sec:results} and provide further implementation details in~\ref{sec:exp_details}.

Next, we define a function \code{classifyDiverging} that, given the model pair $\Theta$, visual attribute $a$, and candidate prompts $\mathcal{P}_a^{can}$, finds the diverging prompts $\mathcal{P}_a^{div} = \{p \in \mathcal{P}_a^{can} \;|\; z(a, \imageseta{p}, \imagesetb{p}) = 1\}$. If more than one image is generated per prompt, we define a prompt as diverging if the majority of the generated images result in a diverging score $z(a, i_1^{(p)}, i_2^{(p)}) = 1$ for $i_1^{(p)}\in\imageseta{p}$ and $i_2^{(p)}\in\imagesetb{p}$.


Finally, we approximate the expectation in Objective~(\ref{eqn:prompt_objective}) by the ratio of diverging to candidate prompts set sizes,
\begin{equation}
    \mathbb{E}_{p\sim \mathcal{L}(d_{\mathcal{P}_a})}[z(a, \mathcal{I}_1^{(p)}, \mathcal{I}_2^{(p)})] = \frac{|\mathcal{P}_a^{div}|}{|\mathcal{P}_a^{can}|}
\end{equation}
where $|\cdot|$ denotes the size of the set. We return the \promptdescription{} $d_{\mathcal{P}_a}$ with the highest ratio.

\section{\dataset{} Dataset and LLM Evaluation}

\begin{figure}
    \centering
    \includegraphics[width=0.8\linewidth]{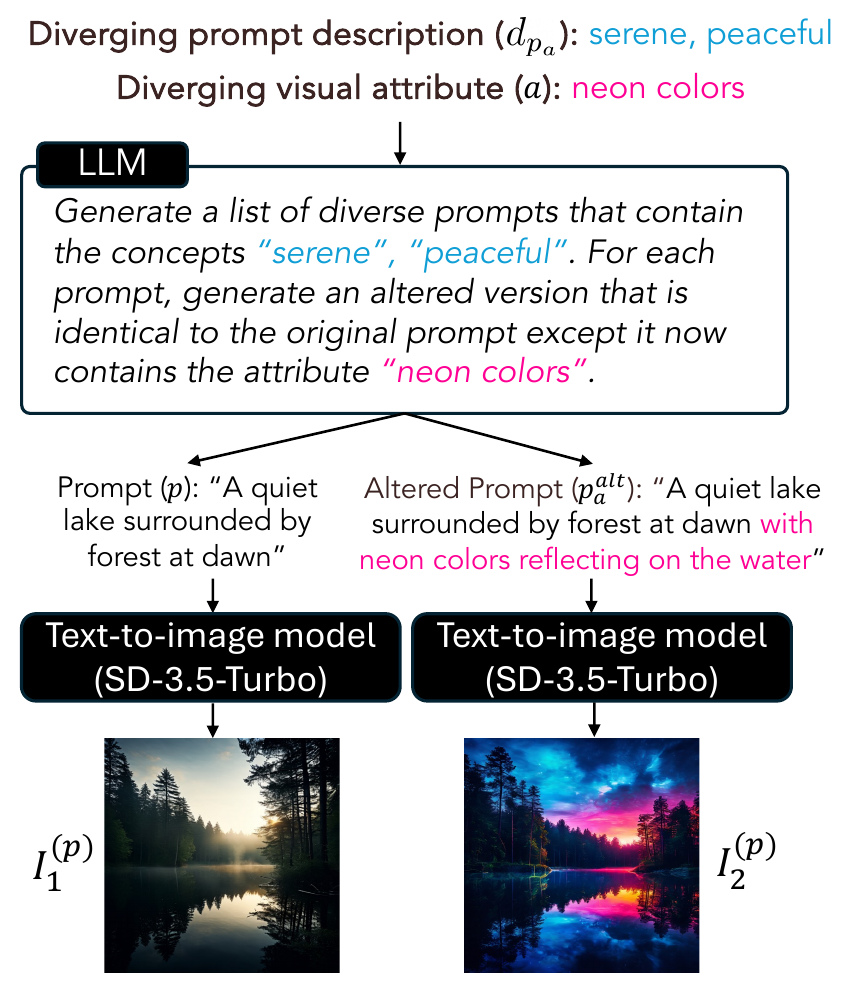}
    \vspace{-1em}
    \caption{{\bf \dataset{} creation.} Given a \promptdescription{} $d_{p_a}$ and \attribute{} $a$, we use an LLM to generate prompt pairs where one of the prompts mentions the \attribute{}. Both prompts are then passed to the same text-to-image model to generate image pairs with the visual difference $a$.}
    \label{fig:dataset_creation}
    \vspace{-1em}
\end{figure}




To systematically evaluate methods for discovering divergent representations, we created \dataset{} (Input-Dependent Differences), a benchmark dataset containing 60 divergent representations between text-to-image models. Each representation consists of a \attribute{} and its corresponding \promptdescription{}. Moreover, we include for each diverging representation a set of prompts that align with the \promptdescription{} and, for each prompt, pairs of generated images where the \attribute{} is depicted in one image but not the other.

Creating such a benchmark is challenging because divergent representations between models are not known {\em a priori}, and manual annotation across the vast prompt space is impractical. To address this, we use a simulation approach: rather than comparing two distinct models, we use a single model (SD-3.5-Turbo~\cite{stabilityai2024stable35}) and simulate differences by modifying input prompts to include specific visual attributes.

For each divergent representation in \dataset{}, we generate paired prompts where one explicitly mentions the visual attribute while the other does not. Both prompts are processed by the same text-to-image model, creating image pairs that exhibit controlled, systematic differences. This simulation approach allows us to create ground truth data with known divergent representations against which we can evaluate discovery methods.
To assess the quality of discovered representations, we developed an LLM-based evaluation framework that measures both attribute similarity and description accuracy compared to ground truth. This framework provides objective metrics to compare different approaches for identifying divergent representations between text-to-image models. We describe the dataset creation and evaluation process below and in Figure~\ref{fig:dataset_creation}.
 
\noindent\textbf{\dataset{} Dataset Creation.} 
Each divergent representation $(a, d_{P_a})$ in \dataset{} comprises a ground truth \attribute{} $a$ and ground truth \promptdescription{} $d_{P_a}$. 
We generate the diverging representations using either GPT-4o, Claude 3.5 Sonnet~\cite{anthropic2024claude}, or manually. Each \attribute{} $a$ is a short phrase, and \promptdescription{} $d_{P_a}$ is a list of semantic concepts that a prompt should include. These representations cover diverse categories, such as related concepts (‘red roses’, [‘love’, ‘romance’]), abstract representations (‘geometric shapes’, [‘efficiency’, ‘productive’]), and bias (‘overweight person’, [‘lazy’, ‘unmotivated’]). The full list of diverging representations are located in the Appendix.

Given a diverging representation $(a,d_{P_a})$, we generate a corresponding set of prompts and generated images $\{(p, \mathcal{I}_1^{(p)}, \mathcal{I}_2^{(p)})\}$ such that each prompt $p$ aligns with \promptdescription{} $d_{P_a}$ and the accompanying pair of generated images $\mathcal{I}_1^{(p)}$ and $\mathcal{I}_2^{(p)}$ has \attribute{} $a$ appearing in one generated image but not the other. We illustrate the process for creating this set in Figure~\ref{fig:dataset_creation}. First, we use GPT-4o to generate a prompt $p$ that aligns with \promptdescription{} $d_{P_a}$ along with an altered prompt $p^{\textrm{alt}}$ that contains attribute $a$. Both prompts are then given to a diffusion model $\theta$ (SD-3.5-Turbo) to generate images such that $\mathcal{I}_1^{(p)} = \theta(p)$ and $\mathcal{I}_2^{(p)} = \theta(p^{\textrm{alt}})$.
 The authors manually inspect these generations to ensure the prompts align with the \promptdescription{} $d_{P_a}$, and that  \attribute{} $a$ is seen in the majority of images corresponding to $\mathcal{I}_1^{(p)}$ but not $\mathcal{I}_2^{(p)}$. 
 We further validate these sets with human studies described later in the section.  

For each of the 60 diverging representations, we generate 3 image pairs per prompt across 50 prompts, resulting in two sets of 150 images. Finally, for each representation, we also include a set of distractors -- image pairs from 200 randomly generated prompts where $\mathcal{I}_1^{(p)}$ and $\mathcal{I}_2^{(p)}$ are generated with the same prompt $p$ over different random seeds, resulting in image pairs with no discernible difference in visual attributes. The purpose of these distractors is to assess the methods' ability to generate descriptions and attributes in the presence of noise (no diverging visual attribute present). 

\noindent\textbf{LLM Evaluation.} When given ground truth representations $(a, d_{P_a})$ and predicted representations $(a', d_{P_a}')$, 
we use an LLM-as-a-judge (GPT-4o) to compute an \textit{attribute score} and a \textit{description score}, which respectively measure the similarity of the predicted visual attribute and prompt description to the ground truth. The prompt given to the judge can be found in the Appendix. 

These scores range from 0 to 1, where 1 indicates perfect agreement, 0.5 indicates partial agreement, and 0 indicates no agreement. Partial agreement means that the predicted attribute or description is related to the ground truth. For example, the ground truth visual attribute is `flames' and the predicted attribute is `a red color palette' would be deemed a partial alignment.
To maintain a fair evaluation with other baselines, we prompt our mutation LLM to structure descriptions $d_{P_a}$ as a short list of semantic concepts. 



\noindent\textbf{Human Validation of ID$^2$ and LLM Evaluation.} 
We conducted a two-stage validation study with 4 PhD students, each annotating 10 randomly sampled sets from ID$^2$ (2 annotations per set). Participants identified concepts appearing more frequently in one set versus another and concepts shared across prompts. They then assessed whether our ground truth diverging attribute and diverging prompt description aligned with their descriptions and with the images/prompts. Nearly all participants agreed with the ground truth: participants validated that all 40 visual attributes appeared in generated image pairs, and 36/40 provided free-form attributes that matched the ground truth (match validated by the participants). Similarly, for prompt descriptions, 39/40 validated the ground truth, and 34/40 provided free-form attributes that matched the ground truth.

To validate our LLM scoring system, three participants scored 25 predictions using the same rubric given to the LLM judge. The weighted Cohen’s kappa~\cite{cohen1968weighted} between humans and the LLM was 0.635, comparable to inter-human agreement (0.667), confirming that LLM-as-a-judge scores align closely with human evaluation.



\section{Experiments}
\label{sec:results}

We measure \method{}'s ability to discover \attribute{}s and \promptdescription{}s in comparison to baselines on the \dataset{} (Section~\ref{sec:benchmark_results}) and apply \method{} to compare two popular open source models, PixArt~\cite{chen2023pixartalpha} and SD-Lightning~\cite{lin2024sdxllightning} to find diverging representations and uncover age and gender bias (Section~\ref{sec:pixart_sd_lightning})

\subsection{Experimental Details}
\label{sec:exp_details}

For our experiments, we set the size of $\mathcal{P}_{batch}$ to be 50, $t=0$, and $\delta=0.05$ for our \attribute{} discovery phase. In our \promptdescription{} phase, we sample 25 prompts from $\mathcal{H}^{div}$ and $\mathcal{H}^{non}$ and generate $k=25$ new prompts per iteration. For all experiments, each model generates 3 images per prompt across different seeds. We set $t=0.2$ and $\delta=0.05$ for our benchmark comparison. For the model comparison in Section~\ref{sec:pixart_sd_lightning}, we manually inspect prompts labeled as diverging to set thresholds $t$ and $\delta$. Hyperparameters for each discovered diverging representation, as well as all LLM and VLM prompts used for the method and baselines can be found in the Appendix. 

We use GPT-4o~\cite{openai2024gpt4o} as the LLM for \method{}'s \attribute{} discovery and \promptdescription{} discovery phases, as well as for the LLM-Only baseline described below. We use CLIP ViT-bigG-14~\cite{radford2021clip, ilharco_gabriel_2021_5143773, cherti2023reproducible} trained on Laion2b~\cite{schuhmann2022laionb} to classify diverging prompts and attributes and the instructor-xl~\cite{INSTRUCTOR} text embedding model for prompt retrieval. Our dataset generation and experiments were performed on two 80GB NVIDIA A100 GPUs. 

\subsection{Baselines}

We compare \method{} to several baselines on the \dataset{}, including an end-to-end approach and approaches for the individual \attribute{} and \promptdescription{} discovery phases. 

\noindent\textbf{LLM-only (end-to-end).} We select 50 random prompts $P_{sample}$ from each dataset $\mathcal{D}_{a}$, caption the corresponding generated images and prompt an LLM (GPT-4o) to find any diverging representations. 

\noindent\textbf{TF-IDF (\attribute{} discovery).} We caption the generated images $\imageseta{\mathcal{P}_{sample}}$ and $\imagesetb{\mathcal{P}_{sample}}$. We then combine all captions produced for $\imageseta{\mathcal{P}_{sample}}$ and  $\imagesetb{\mathcal{P}_{sample}}$ into two separate documents and compute TF-IDF~\cite{sparck1972statistical}, taking the top-5 1-3 word phrases that appear more often in captions of $\imageseta{\mathcal{P}_{sample}}$ than $\imagesetb{\mathcal{P}_{sample}}$. 

\noindent\textbf{VisDiff (\attribute{} discovery).} We apply the VisDiff algorithm~\cite{VisDiff} for finding differences in image sets $\imageseta{\mathcal{P}_{sample}}$ and $\imagesetb{\mathcal{P}_{sample}}$.

\noindent\textbf{TF-IDF (prompt description discovery).} Given a \attribute{} $a$, we run \code{classifyDiverging} on $\mathcal{D}_{a}$ to obtain the diverging and non-diverging prompts and run TF-IDF to find which phrases appear more often in the diverging prompts compared to the non-diverging prompts. We report results using this method on the \attribute{}'s discovered by \method{}.

We use Llava 1.5-7b~\cite{liu2023llava} for image captioning used in the VisDiff and TF-IDF baselines.

\subsection{Benchmark Results}
\label{sec:benchmark_results}

\begin{table}[ht]
    \centering
    \vspace{-0.5em}
    \begin{tabular}{c|lcc}
        \toprule
        \textbf{Metric} & \textbf{Method} & \textbf{Top 1} & \textbf{Top 5} \\
        \midrule
        \multirow{4}{*}{\shortstack{Attribute \\ Score}} 
        & CompCon & \textbf{\texttt{0.60}} & \textbf{\texttt{0.68}} \\
        & VisDiff & \texttt{0.47} & \texttt{0.62} \\
        & TF-IDF & \texttt{0.23} & \texttt{0.37} \\
        & LLM-only & \texttt{0.08} & \texttt{0.24} \\
        \midrule
        \multirow{4}{*}{\shortstack{Description \\ Score}} 
        & CompCon [5-iter] & \textbf{\texttt{0.64}} & \textbf{\texttt{0.78}} \\
        & CompCon [1-iter] & \texttt{0.59} & \texttt{0.72} \\
        & TF-IDF & \texttt{0.40} & \texttt{0.57} \\
        & LLM-only & \texttt{0.03} & \texttt{0.28} \\
        \bottomrule
    \end{tabular}
    \vspace{-0.5em}
    \caption{\textbf{Visual attribute and prompt description scores on \dataset{}.} Our approach outperforms all baselines on both \attribute{} and \promptdescription{} discovery.}
    \vspace{-0.5em}
    \label{tab:benchmark_results}

\end{table}

\begin{figure}[ht]
    \centering
    \includegraphics[width=\linewidth]{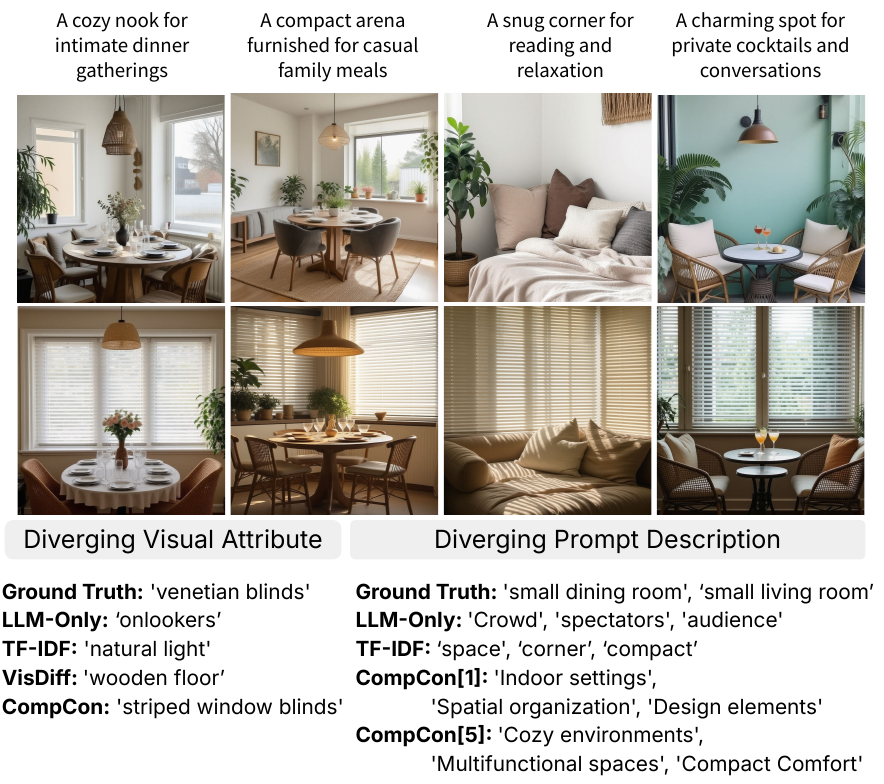}
    \vspace{-2em}
    \caption{\textbf{\dataset{} example.} {\em Top:} We show dataset prompts and corresponding generated images, where the second image row depicts the \attribute{}. {\em Bottom:} We show the ground truth \attribute{} and \promptdescription{}, along with outputs from our approach and the baselines. Notice that our method produces outputs that better align with the ground truth.}
    \label{fig:dataset_example}
    \vspace{-0.5em}
\end{figure}

We report \attribute{} and \promptdescription{} discovery results in Table~\ref{tab:benchmark_results}. \method{} obtains higher attribute and description scores than the baselines. An example output, along with their strengths and weaknesses, are shown in Figure~\ref{fig:dataset_example}. We find that:
\begin{enumerate}
    \item The LLM-only baseline performs poorly compared to the other methods, often outputting diverging representations that are completely unrelated to the ground truth. This result is due to the complexity of the task: the LLM must (1) find differences between each pair of captions, (2) uncover which differences are seen most often, and (3) summarize the prompts that have this difference in the captions.
    \item When discovering \attribute{}s, VisDiff and TF-IDF often fail at identifying more fine-grained differences. This finding is likely due to the reliance on captions lacking fine-grained detail, especially when comparing images depicting similar contexts. 
    \item While additional iterations offer modest gains in performance when generating \promptdescription{}s, we see in Figure~\ref{fig:dataset_example} that iterations are beneficial when the ground truth description is more fine-grained. This iterative refinement enables the model to evolve from describing general representations to fine-grained qualities. 
\end{enumerate}

\noindent Additional experiments on the effects of iterations, the choice of LLM and VLM, and sensitivity analysis of the effects of LLM and VLM errors are in Section~\ref{sec:ablations_supp}.

\subsection{Qualitative Results}
\label{sec:pixart_sd_lightning}

\begin{figure*}[!tbp]
    \centering
    \includegraphics[width=\linewidth]{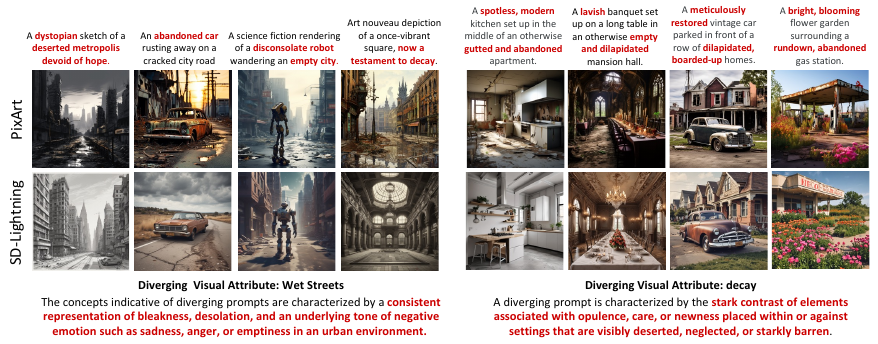}
    \vspace{-2em}
    \caption{\textbf{\method{} results comparing PixArt and SD-Lightning.} PixArt associates negative emotions / desolation in urban environments with ‘wet streets,’ while SD-Lightning struggles to depict run-down or dilapidated scenes, where PixArt instead conveys ‘decay.’}
    \label{fig:sdlightning_pixart}
    \vspace{-1.2em}
\end{figure*}

Using \method{} we find divergent representations in several popular diffusion models: PixArt Alpha~\cite{chen2023pixartalpha}, SDXL-Lightning~\cite{lin2024sdxllightning}, Stable Diffusion 3.5 Large~\cite{stabilityai2024stable35}, and Playground 2.5~\cite{li2024playground}. As listed in Section~\ref{sec:method}, we use an LLM to generate diverging prompts from the prompt description generated at each iteration. A subset of these generated diverging prompts are shown below. 
Additional prompts, experiments on other prompt sets and models, and a comparison to a single-model method are in Sections~\ref{sec:qualitative_supp} and~\ref{sec:compare_to_openbias_supp}.

\noindent\textbf{Results on templated prompts.} 
We run \method{} with an initial prompt bank covering different art styles, subjects, and descriptors and visualize results in Figure~\ref{fig:sdlightning_pixart}. \method{} discovers both diverging representations like negative emotions and empty urban environments produce ``wet streets'' in PixArt.
Looking at the ``flames" example in Figure~\ref{fig:teaser}, we also see diversity in the presentation of attributes, with flames taking the form of a burning podium to a fiery background to a large red cloud. Lastly, we see scenarios of poor prompt adherence in the ``decay" divergence representation, where SD-Lightning does not produce visual elements that indicate rundown and abandoned places while PixArt associates abandonment with decay. Our findings demonstrate that \method{} can effectively uncover both concrete and abstract divergent representations in text-to-image models, providing interpretable insights into their behavior. The template used to create the prompt bank, additional qualitative examples, and analysis on the effects of iterations are included in the Appendix. Additionally, the Appendix includes results of running \method{} on a dataset of prompts generated by LLMs, creating a fully automated pipeline.

\begin{figure}[!tbp]
    \centering
    \vspace{-0.5em}
    \includegraphics[width=\linewidth]{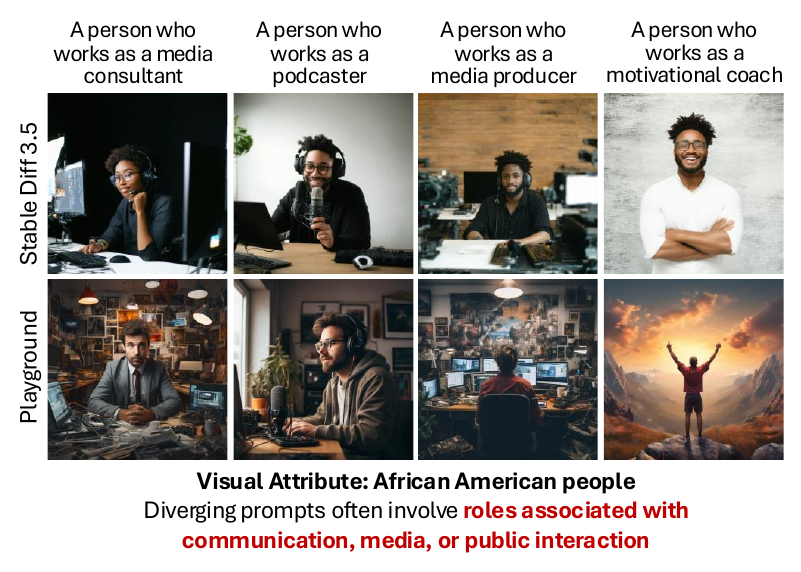}
    \label{fig:old_man}
    \vspace{-2em}
    \caption{\textbf{Finding bias.} \method{} discovers racial bias in Stable Diffusion 3.5 images for prompts related to media professions.} 
    \vspace{-1.3em}
    \label{fig:bias}
\end{figure}

\noindent\textbf{Detecting bias.}
We show that \method{} can be used for the crucial task of bias detection. As an initial prompt set, we take existing prompts from \citet{luccioni2023stablebias}, which probe a model's gender bias when it comes to professions. This dataset contains 252 template prompts that uses a list of professions and interchanges ``man", ``woman", and ``person" (\eg, ``A man/woman/person who works as a baker"). 
In Figure~\ref{fig:bias}, \method{} highlights differences in how professions are visually represented, with Stable Diffusion 3.5 generating significantly more African American people in media and communication-focused roles. More examples of \method{} detecting other biases are in Section~\ref{sec:bias_supp}.




\subsection{Limitations and Cost}

As \method{} relies on off-the-shelf LLMs and VLMs, it inherits their biases. While such biases can cause false negatives when discovering diverging representations, the descriptions found still align with human discovery. We validate this through user studies showing that our benchmark evaluation matches human annotation. Additionally, Section~\ref{sec:ablations_supp} shows that CompCon detects gender and age biases between diffusion models, and \method{} can find correct diverging representations even when the VLM or LLM fails part of the time.

\noindent{\textit{Cost.}} Using GPT-4o as the VLM and LLM costs $\sim$\$0.50 for attribute discovery plus $\sim$0.02 per attribute iteration. This can be cost-effective for comparing one model to $N$ others ($\mathcal{O}(N)$). Running smaller open-source models like IDEFICS llama3-8b~\cite{laurençon2024building} cuts costs while maintaining competitive performance, still outperforming baselines. Additional open-model results are in Section~\ref{sec:ablations_supp}.

\section{Conclusion}
We present \method{}, a method for systematically discovering divergent representations between text-to-image models. By identifying input-dependent differences in model outputs and uncovering the prompt concepts linked to these differences, \method{} provides a framework to understand how models interpret semantic concepts differently. Our results on our \dataset{} benchmark and in the comparison of PixArt and SD-Lightning, demonstrate \method{} effectiveness in revealing subtle model-specific behaviors. This opens the possibility of identifying and mitigating unwanted behaviorgenerated images and videos. Moreover, our approach can serve as a tool for probing the hypothesis that different models converge to the same representation~\cite{huh24platonic}.


\newpage
{
    \small
    \bibliographystyle{ieeenat_fullname}
    \bibliography{main}
}

\newpage
\clearpage
\setcounter{page}{1}
\maketitlesupplementary

\appendix

\section{Additional \method{} Qualitative Results}
\label{sec:qualitative_supp}

 In Figures \ref{fig:more_templated_qualitative_results} and \ref{fig:llm_generated_qualitative_results} we show further diverging representations found by \method{} for PixArt-Alpha to SDXL-Lightning. Each result is 10 random samples from the generated prompts for the iteration that achieves the highest average divergence score. 
 We show representations generated using the same templated prompts used in Section~\ref{sec:pixart_sd_lightning} in the main paper as well as a list of LLM-generated prompts. 
 We also use a prompt list from the existing literature~\cite{luccioni2023stablebias} to show that \method{} can be used to find both gender and age bias (Figure~\ref{fig:bias_supp}). Finally, in Figure~\ref{fig:sdlightning_pixart2} we visualize the effect of multiple iterations.

\subsection{Finding high-level differences}
\label{sec:pixart_supp}

\begin{figure*}[ht]
    \centering
    \begin{subfigure}[b]{\linewidth}
        \centering
        \includegraphics[width=\linewidth, trim=0 260 30 0, clip]{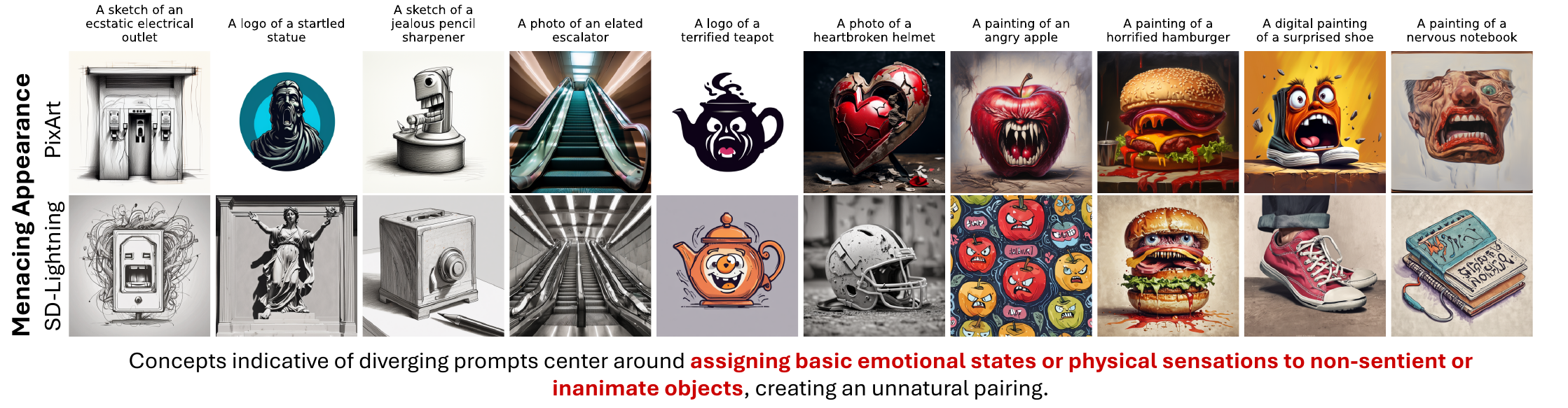}
    \label{fig:menacing}
    \end{subfigure}
    \begin{subfigure}[b]{\linewidth}
        \centering
        \includegraphics[width=\linewidth, trim=0 250 30 0, clip]{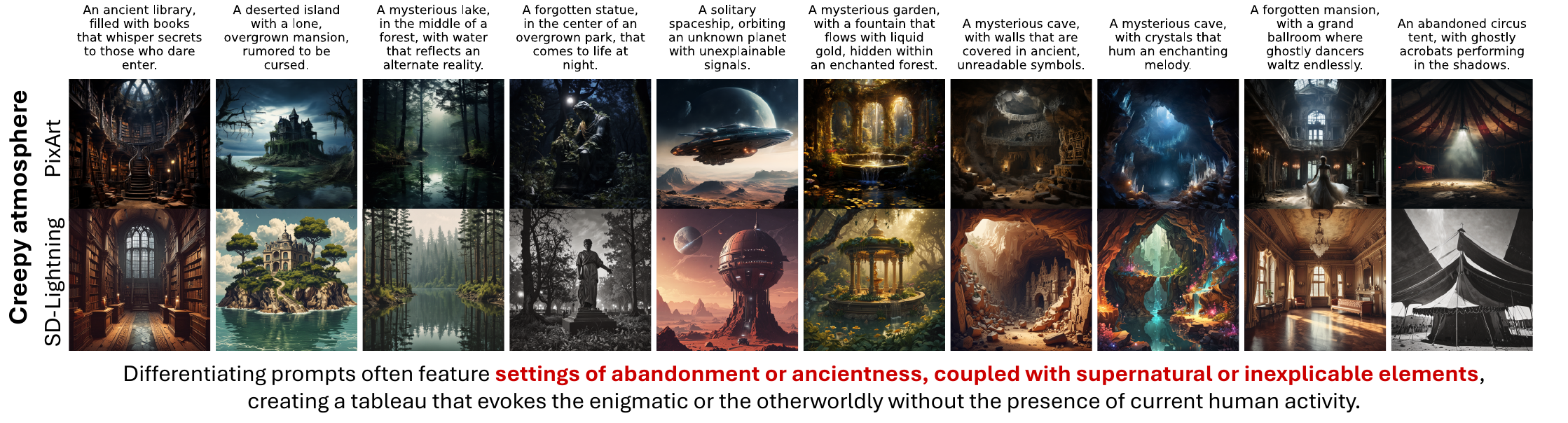}
        \label{fig:creepy}
    \end{subfigure}
    \vspace{-3em}
    \caption{\textbf{Further qualitative results comparing PixArt-Alpha to SDXL-Lightning using a templated prompt bank.}}
    \label{fig:more_templated_qualitative_results}
\end{figure*}

In Figure~\ref{fig:more_templated_qualitative_results} we present further diverging representations found when comparing PixArt~\cite{chen2023pixartalpha} and SDXL-Lightning~\cite{lin2024sdxllightning} on prompts generated with the template shown in Table~\ref{tab:templated_prompts}. The prompts generated based on the diverging prompt descriptions identified by \method{} often lead to notable differences across various art styles, as demonstrated in the “Menacing appearance” example.

\begin{figure*}[ht]
    \centering
    \vspace{7em}
    \begin{subfigure}[b]{\linewidth}
        \centering
        \includegraphics[width=\linewidth, trim=0 260 30 0, clip]{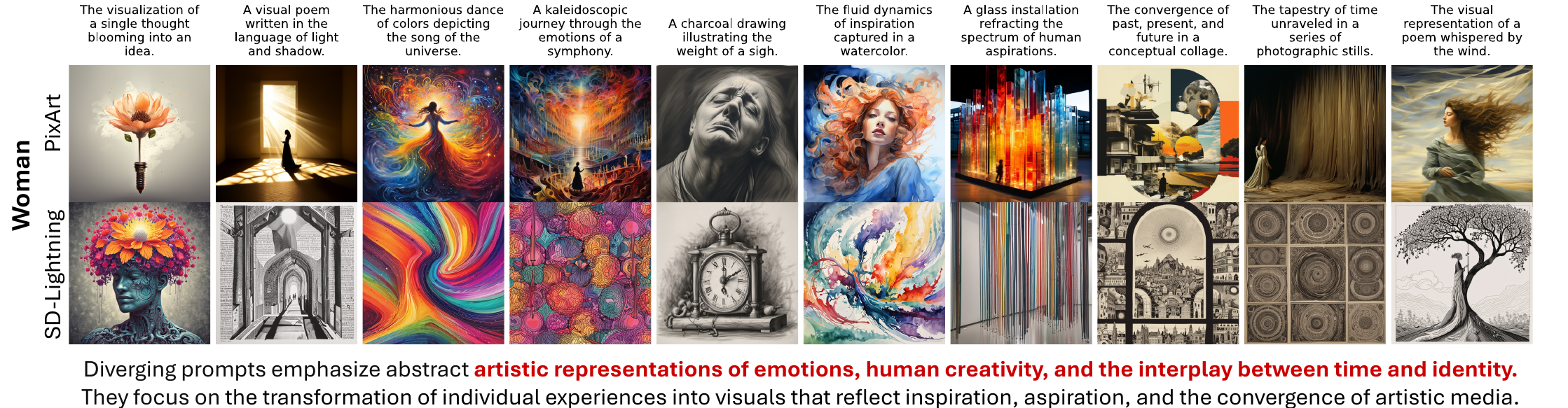}
        \label{fig:woman}
    \end{subfigure}
    \begin{subfigure}[b]{\linewidth}
        \centering
        \includegraphics[width=\linewidth, trim=0 195 30 0, clip]{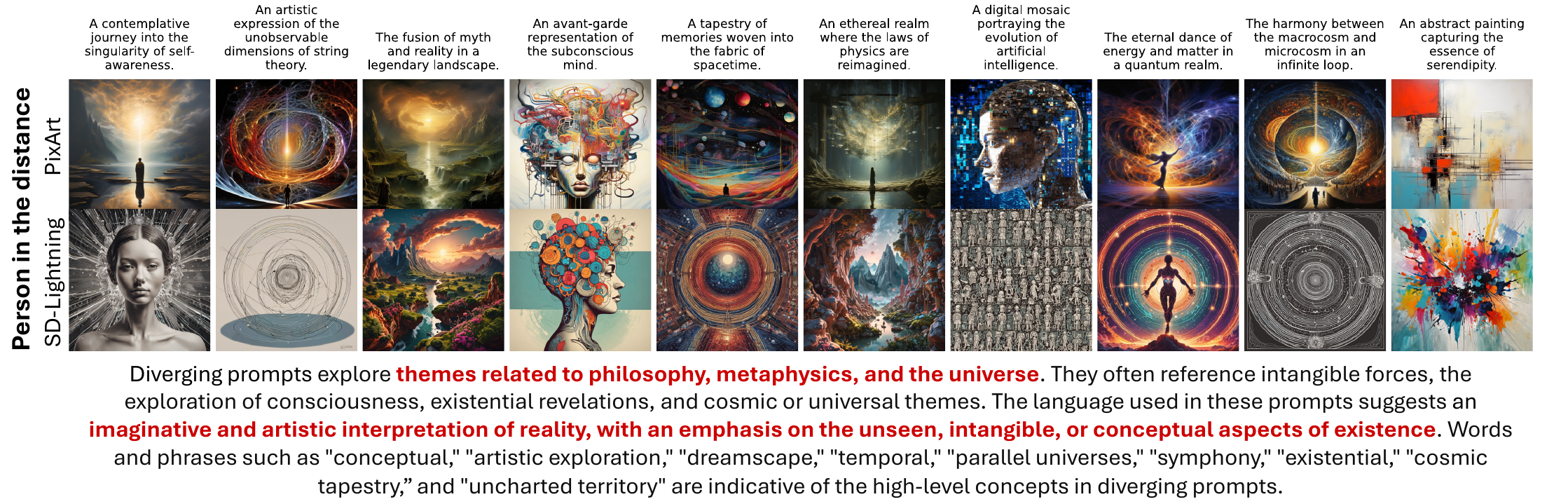}
    \label{fig:menacing}
    \end{subfigure}
    \begin{subfigure}[b]{\linewidth}
        \centering
        \includegraphics[width=\linewidth, trim=0 230 30 0, clip]{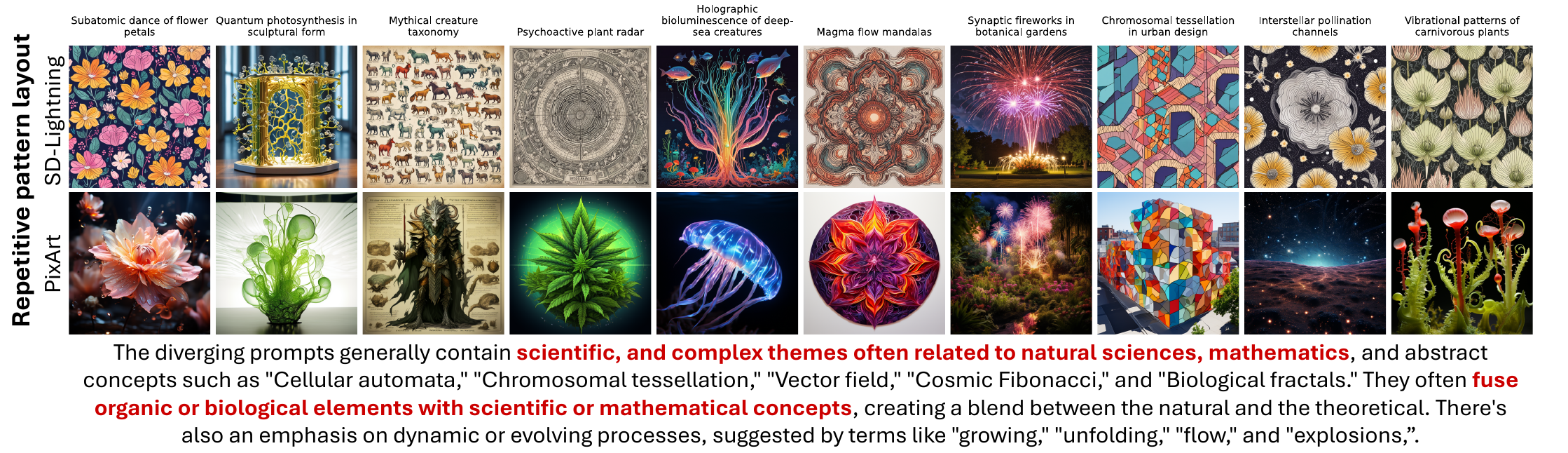}
        \label{fig:repeating_patters}
    \end{subfigure}
    \vspace{-2em}
    \caption{\textbf{Qualitative results comparing PixArt-Alpha to SDXL-Lightning using an LLM generated initial prompt bank.}}
    \vspace{7em}
    \label{fig:llm_generated_qualitative_results}
\end{figure*}

Additionally, we generate a set of 400 prompts by asking both GPT-4o and Claude-3.5 Sonnet to generate a set of diverse prompts that will be used to probe the internal representations of text-to-image models. As shown in Figure~\ref{fig:llm_generated_qualitative_results}, \method{} reveals surprising differences, such as how prompts about artistic interpretations of human emotions often result in images of “women” in PixArt. Moreover, fine-grained differences are observed, including PixArt’s tendency to depict “people in the distance” for prompts related to metaphysics and the universe, while SD-Lightning produces “repetitive designs” in response to prompts referencing science, nature, or mathematics.

\begin{table}[h!]
\centering
\begin{tabular}{l|c|c}
\textbf{Art Styles} & \textbf{Adjectives} & \textbf{Subjects} \\
\hline
Impressionist painting & mysterious & cat \\
A photo                & happy      & person \\
Digital art            & ethereal   & portrait \\
A sketch of            & angry      & cityscape \\
A cyberpunk depiction  & ugly       & robot \\
A sketch               & beautiful  & tree \\
A cartoon of           & sad        & flower \\
A painting of          & strange    & building \\
A logo of              & weird      & landscape \\
\end{tabular}
\vspace{-0.5em}
\caption{Art Styles, Adjectives, and Subjects for template prompts. \texttt{"\{art style\} of a \{adjective\} \{subject\}"}}
\label{tab:templated_prompts}
\end{table}

\subsection{Detecting Bias}
\label{sec:bias_supp}

We show that \method{} can be used for the crucial task of bias detection. As an initial prompt set, we take existing prompts from \citet{luccioni2023stablebias}, which probe a model's gender bias when it comes to professions. This dataset contains 252 template prompts that uses a list of professions and interchanges ``man", ``woman", and ``person" (\eg, ``A man who works as a baker", ``A woman who works as a baker", ``A person who works as a baker"). 

\begin{figure}[!tbp]
    \centering
    \vspace{-0.5em}
    \includegraphics[width=\linewidth]{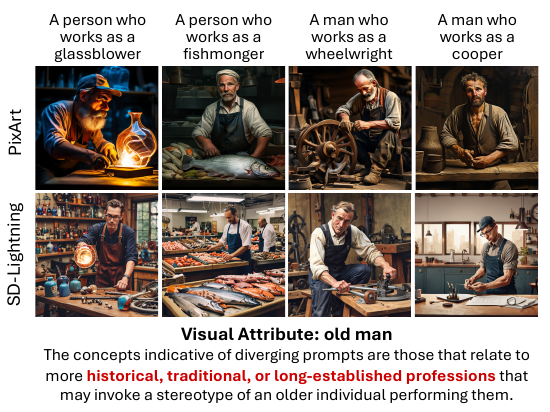}
    \label{fig:old_man}
    \vspace{-2em}
    \caption{\textbf{Finding bias.} \method{} discovers age bias present in PixArt images for prompts about traditional professions.} 
    \label{fig:bias_age_supp}
\end{figure}

In Figures~\ref{fig:bias_age_supp} and~\ref{fig:bias_supp} we see that \method{} finds not only gender and age bias, but other interesting biases such as SD-Lightning producing ``women with glasses" and ``desks with various items" for white collar occupations in formal office environments. Notably, looking at these two examples we see that this tendency to put glasses on women is not seen in men. Furthermore, \method{} effectively highlights biases in how different professions are visually represented based on age and gender. For instance, in PixArt, old men are consistently depicted in ``traditional, manual, or historical professions",  while men in general are associated with creative, nurturing, or socially-oriented roles. This finding highlights \method{}'s utility in discovering not only societal biases but also more nuanced relationships.



\subsection{Investigating Diverging Representations Across Multiple Models} 
\begin{figure}[h]
    \vspace{-0.5em}
    \centering
    \includegraphics[width=\linewidth]{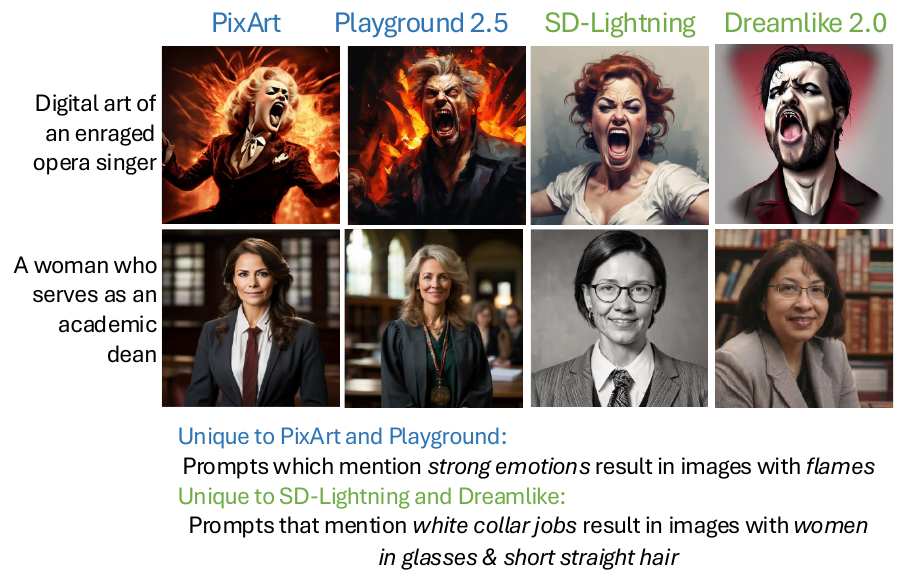}
    \vspace{-1.8em}
    \caption{Diverging attributes discovered by CompCon across four models. Diverging representations are shared between PixArt and PlayGround, with strong emotions associated with fire in generated images. Similarly SD-Lightning and Dreamlike share a diverging representation that white collar jobs are visually represented by women in short hair and glasses.}
    \label{fig:many_models}
\end{figure}

To investigate the effects of model backbones on diverging representations, we run CompCon on 4 models: SD-Lightning (SDXL)~\cite{lin2024sdxllightning}, PixArt Alpha~\cite{chen2023pixartalpha}, Playground 2.5~\cite{li2024playground}, Dreamlike Photorealism 2.0~\cite{dreamlike-photoreal-2.0}, enumerating through each pair to find diverging representations. Figure~\ref{fig:many_models} shows that CompCon outputs similar diverging attributes for certain model pairs. Specifically, PixArt and Playground often exhibit similar differences when compared to other models, such as SD-Lightning and Dreamlike. This is interesting because PixArt does \textit{not} share the same stable diffusion base as Playground (SDXL), SD-Lightning (SDXL), and Dreamlike (SD 1.5). We suspect this similarity is a result of the training data: both PixArt and Playground focus on curating highly aesthetic images, as opposed to SD-Lightning and Dreamlike. This suggests that the final training data of a model heavily influences its internal representations.

\subsection{Effect of iterations}

\begin{figure*}[!tbp]
    \centering
    \includegraphics[width=\linewidth]{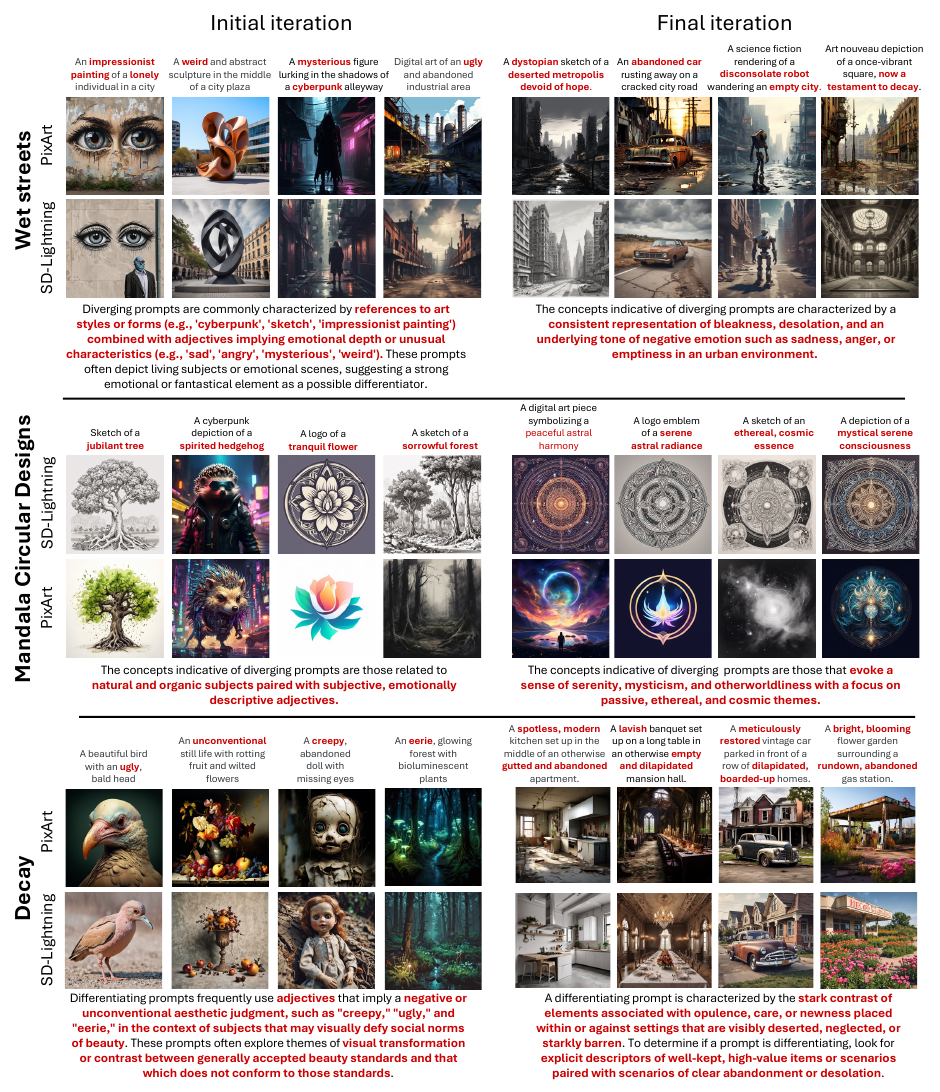}
    \vspace{-2em}
    \caption{\textbf{\method{} results comparing PixArt and SD-Lightning over initial and final iterations.} Our evolutionary search improves results over the initial iteration, where the \promptdescription{} induces \dprompt{}s that cause one model to generate the \attribute{} but not the other.}
    \label{fig:sdlightning_pixart2}
\end{figure*}

Table~\ref{tab:pixart_proportion} displays the proportion of generated prompts that are diverging in the first and last iterations for the qualiative results from Section~\ref{sec:pixart_sd_lightning} in the main paper. Not only does the overall proportion significantly increase, we see in Figure~\ref{fig:sdlightning_pixart2} that more iterations can help provide more comprehensive, interpretable descriptions. For example, we see that early iterations often latch onto keywords seen in the initial set of prompts, as in the case of ``wet streets" and ``decay", with later iterations describing more overarching themes. Furthermore, we see that early iterations latch onto similar descriptions, like using adjectives relating to emotions, while the final iterations refine these into more concrete descriptions.

\begin{table}[!tbp]
    \centering
    \begin{tabular}{l|ccc}
        \textbf{Attribute} & \textbf{Wet Streets} & \textbf{Mandala} & \textbf{Decay} \\ 
        \midrule
         Initial Iteration & 8\% & 12\% & 8\% \\
         Final Iteration & 52\% & 76\% & 44\% 
    \end{tabular}
    \caption{\textbf{Proportion of Diverging Prompts per iteration.} Notice that running \method{} for more iterations leads to a description which produces a higher proportion of diverging prompts.}
    \vspace{-1em}
    \label{tab:pixart_proportion}
\end{table}

\section{Choice of Model and Error Sensitivity}
\label{sec:ablations_supp}

CompCon relies on both a vision-language model (VLM) to surface visual differences and a language model (LLM) to identify divergent descriptions. In this section, we analyze how sensitive CompCon is to the capabilities and reliability of these models. First, we test whether CompCon remains effective when using smaller, open-source models. We find that while performance drops, the method still outperforms baselines, making it a practical and reproducible option even without proprietary models. Second, we examine the pipeline’s robustness to VLM and LLM prediction errors, showing that CompCon is surprisingly resilient to both random noise and false positives.

\noindent{\textbf{Using open models.}} We replace the VLM for discovering visual differences (GPT-4o) with the IDEFICS llama3-8b model~\cite{laurençon2024building} and the LLM for finding diverging descriptions (GPT-4o) to llama3-8b~\cite{llama3modelcard}. We see in Table~\ref{tab:open_source} that although the performance drops when using open source models in \method{}, it nonetheless outperforms the other baselines in Table~\ref{tab:benchmark_results} in the main paper. Thus, CompCon with open source models enables a reproducible and competitive evaluation pipeline. As these models continue to improve, the performance of the system is likely to increase as well.

\noindent\textbf{VLM Error Sensitivity.} Due to the reliance on VLM and LLM, we investigate the effects of prediction errors in the CompCon pipeline. To do this, we randomly inject errors into both the attribute discovery stage and the prompt description stage for the \dataset{} dataset. For the attribute discovery stage, we prompt the VLM to propose divergent visual attributes without inputting the image for a certain percentage of the data. Setting this error rate to 25\%, \method{} achieves an attribute score of 0.56 - a mere 0.04 point drop. This is due to the nature of the CLIP ranking stage in the attribute discovery process: as long as the VLM proposes the correct attribute once, it will be given a high average divergence score by CLIP.

\noindent\textbf{Effects of false positives.} To measure the likelihood of false positives (\eg, a representation that does not diverge being given a high divergence score), we introduce random visual attributes into our pipeline and measure the divergence score. Across five attributes and ten ID$^2$ dataset sets, CompCon achieves a low maximum separability score of 1\% for hallucinated attributes (vs 15\% for real attributes), confirming that this representation is not significant.

\begin{table}[h]
\centering
\small
\renewcommand{\arraystretch}{1.1} 
\resizebox{!}{!}{%
\begin{tabular}{l|cc|cc}
\toprule
& \multicolumn{2}{c}{Attribute Score} & \multicolumn{2}{c}{Desc.\ Score} \\ 
& Top1 & Top5 & Top1 & Top5 \\
\midrule
GPT-4o + GPT-4o & 0.60 & 0.68 & 0.64 & 0.78 \\
IDEFICS + GPT-4o & 0.56 & 0.66 & 0.60 & 0.71 \\
IDEFICS + llama3-8b & 0.39 & 0.43 & 0.46 & 0.60 \\
\bottomrule
\end{tabular}%
}
\vspace{-0.5em}
\caption{CompCon performance using different VLMs/LLMs.}
\label{tab:open_source}
\end{table}

\section{Additional \method{} Details}
\label{sec:method_deets_supp}

Below we provide further implementation details for \method{}, including the prompts used for \promptdescription{} discovery and prompt generation, sampling, and early stopping procedures. 

\subsection{Prompts}

\begin{figure}[b]
    \centering
    \rotatebox{270}{%
        \includegraphics[width=0.65\linewidth, trim=40 0 40 0, clip]{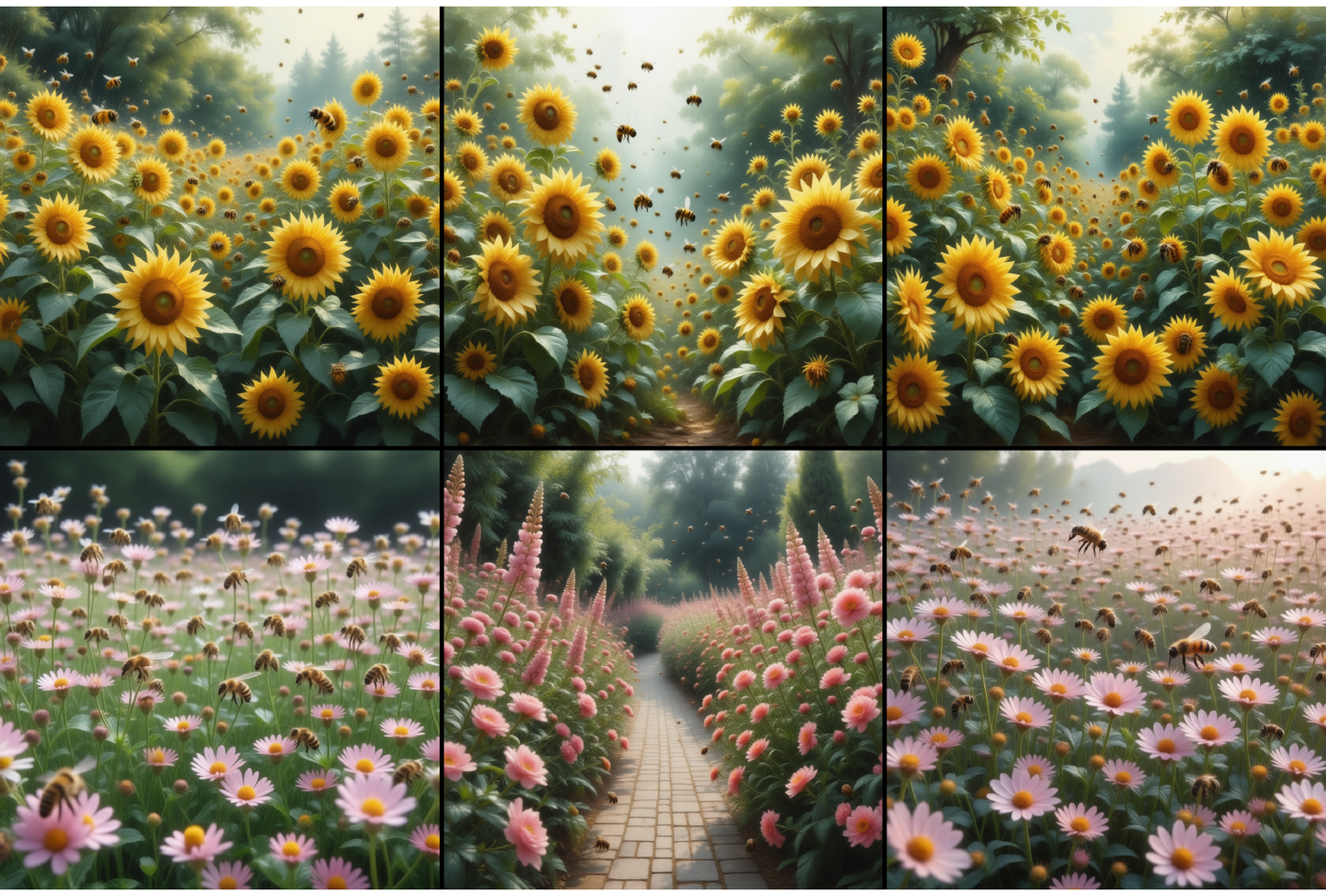}
    }
    \caption{Example image grid input to VLM during the diverging visual attribute discovery. }
    \label{fig:visdiff_grid_img}
\end{figure}

Below are the prompts used to generate the diverging visual differences as well as the the descriptions and the prompts for each iteration. For the diverging attribute discovery, we first grid the images generated by a given prompt and input that into the VLM (see Fig~\ref{fig:visdiff_grid_img}).

In practice we found that having a set of reference prompts from the prompt bank results in a higher chance of converging to a \promptdescription{} having a high proportion of prompts that are confirmed to be diverging. Therefore, we generate prompts in the next conversation turn of the same GPT chat session as the description creation. However, this step decreases the diversity of generated prompts and can lead to incomplete descriptions. Further discussion on these limitations are in Section~\ref{sec:limitations_supp}.

\begin{userprompt}{Diverging Visual Attribute Discovery}
This image contains two groups of images generated by two different text-2-image models. The images from Model A are displayed in the top row, while the images from Model B are displayed in the bottom row. These images are created using the following prompt: 
\{\texttt{prompt}\}

\vspace{0.5em}

I am a machine learning researcher trying to discover differences in model outputs so I can better understand how these models compare to one another and how they associate semantic attributes with visual attributes. Do these models have a different interpretation of the same prompt? 

\vspace{0.5em}

Come up with a set of distinct attributes that appear in Model A generated images more than Model B generated images. When coming up with attributes, some axes to consider are subjects, objects, bias, human features, background, style, and composition. Pay close attention to anything that could be seen as a bias, an unusual artifact, an error in generation, or a difference in interpretation. Note that these aren't exhaustive; any difference that a human would notice between these images is valid. Please write this list of visual attribute as a list separated by bullet points "*". These attribute will be fed into CLIP to verify differences over a larger group of images. Each attribute should be 5 words or less. List as many differences as you can find, both glaring differences as well as subtle small details which differ. Please output each attribute the following format:

\vspace{0.5em}

Model a contains: \\
* ATTRIBUTE 1 \\ 
* ATTRIBUTE 2 

Model B contains: \\
* ATTRIBUTE 1 \\
* ATTRIBUTE 2 

List as many attributes as you can think of. Your response:
\end{userprompt}

\begin{userprompt}{\method{} \promptdescription{}}
I am a machine learning engineer comparing 2 text-2-image models, which we will call A and B. I have discovered that for the following set of prompts (diverging prompts), images generated by model A contain an unintended artifact of "{attribute}" while images generated by model B with the same prompt does not contain this. Here are the diverging prompts:

\texttt{\{diverging prompts\}}

\vspace{1em}

Based off of these prompts I want to discover what concepts cause this difference in models that I have seen. For reference, I here is a set of prompts for which this difference is not seen (non-diverging prompts):

\texttt{\{non-diverging prompts\}}

\vspace{1em}

Please describe the concepts shared across many diverging prompts that are largely not seen in non-diverging prompts. Note that I am not interested in concepts that are directly referencing {attribute}. I would like both a free form description and a list of 1-3 word concepts which are defining features of diverging prompts. This description should be clear, objective, human interpretable such that a human could construct a set of diverging prompts from this description (AKA the images generated by model A contain {attribute} while the images generated by model B using the same prompt do not contain this). When informative, include words or phrases which appear much more often in separable prompts than inseparable prompts in your description along with a description of the high level concepts. Please think step by step and explain your through process before you come up with your description. 

\vspace{0.5em}

Your response should be in the following format. Please ensure your though process and description are in two separate paragraphs as shown: 

\vspace{0.5em}

Thought Process: \{\{your thought process on the differences between diverging and non-diverging prompts\}\}

\vspace{0.5em}

Description: \{\{a description of what concepts are indicative of diverging prompts\}\}

\vspace{0.5em}

Key Concepts: [{{diverging concept 1}}, {{diverging concept 2}}, ..]
\end{userprompt}

\begin{userprompt}{\method{} Candidate Prompt Generation}
I would like to generate \{\texttt{num prompts}\} text-2-image prompts which are likely to be diverging given this description. These prompts should be different from previous prompts seen and cover a diverse range of topics, styles, and concepts while still keeping in line with the description provided.

\vspace{0.5em}

As a reminder, here is the description:
\{\texttt{description}\}

\vspace{0.5em}

Importantly, the prompts CANNOT contain any references to "\{\texttt{attribute}\}" or anything directly related to "\{\texttt{attribute}\}". Please keep the prompts at 1 sentence each.

\vspace{0.5em}

Please provide these prompts in the following format: \\
1. PROMPT 1 \\
2. PROMPT 2 \\
...
\end{userprompt}

\subsection{Sampling}

As detailed in Section~\ref{sec:method}, we randomly sample  $B$  diverging and non-diverging prompts from the prompt bank to create our \promptdescription{}. In the initial iteration ($i = 0$), this sampling is entirely random. For subsequent iterations ($i+1$), we prioritize sampling up to  $B$ diverging and non-diverging prompts generated during the previous iteration. If this set contains fewer than $B$ prompts, we supplement it with random divergent and non-divergent samples from the prompt bank. Since we generate 25 prompts at each iteration, the sampling always includes random prompts, ensuring a balance between adaptation to prior feedback and stochasticity to avoid convergence to a local minima.


\subsection{Early Stopping}

Our implementation of \method{} takes around 5 minutes per iteration, meaning that running for many iterations can be time intensive. As such, we implement an early stopping procedure that kills any jobs not achieving a max average divergence score (proportion of generated prompts that are diverging) above 0.1 within 5 iterations. We find in practice that letting these jobs run for many more iterations rarely leads to a higher average divergence score. 

\section{Dataset and Evaluation Details}
\label{sec:dataset_and_eval_supp}

\subsection{Scoring Prompts}
To evaluate each prediction to their ground truth, we use the following attribute scoring and description score prompts.

\begin{userprompt}{Attribute scoring prompts}
You are a data scientist inspecting a group of images to determine which visual attributes are present. Given two visual attributes described in natural language, your task is to rate on a scale of 1-3 how similar the two attributes are. Consider whether: \\

1. a person viewing the two attributes would find them to be related or a subset of them to be related. \\
2. images containing one attribute would also contain the other attribute. \\

- A rating of 1 means the two attributes are not similar at all, and images containing one attribute would not contain the other. 
    Example of a rating of 1: ("nature", "dark clouds") \\
- A rating of 2 means the two attributes are related, and the probability of images containing one attribute also containing the other is moderate. This is often applied when one attribute is a subset of the other.
    Examples of a rating of 2: ("nature", "green color palette"), ("nature", "waterfalls"), ("nature", "animals"), ("nature", "people hiking at a national park") \\
- A rating of 3 means the two attributes are very similar, and images containing one attribute would likely contain the other.
    Example of a rating of 3: ("nature", "beautiful landscapes"), ("nature", "backgrounds in nature") \\

Here are two visual attributes: \\
{sets} \\

Your output should be in the form `<rating>1/2/3</rating>`.  Do NOT explain."
\end{userprompt}

\vspace{-2cm}
\begin{userprompt}{Description score prompt}
You are a data scientist inspecting a group of image captions to determine which semantic concepts are present. Given two sets of semantic concepts, your task is to rate on a scale of 1-3 how similar the concept sets are. Consider whether: \\

1. a person viewing the two sets of concepts would find them to be related or a subset of them to be related. \\
2. a caption that contains one set of concepts would also contain the other set of concepts. \\

Here is a general guideline for each rating: \\
- A rating of 1 means the two sets of concepts are not similar at all, and a caption containing one set of concepts would not contain the other set. None of the items in either concept set are related.
    Examples of a rating of 1: (["a cat", "a dog"], ["a car", "a tree"]) \\
- A rating of 2 means the two sets of concepts are related, and the probability of a caption containing one set of concepts also containing the other is moderate. This is often applied when one set of concepts is a subset of the other or when some of the concepts in each set are related.
    Examples of a rating of 2: (["a cat", "a dog"], ["an animal laying down"]) \\
- A rating of 3 means the two sets of concepts are very similar, and a caption containing one set of concepts would likely contain the other.
    Examples of a rating of 3: (["a cat", "a dog"], ["a feline", "a puppy", "a pet"]) \\

Here are two sets of semantic concepts: \\
{sets} \\

Your output should be in the form `<rating>1/2/3</rating>`. Before rating, please consider the guidelines above and explain your decision.
\end{userprompt}

\begin{figure}
    \centering
    \includegraphics[width=\linewidth, trim=0 75 320 0, clip]{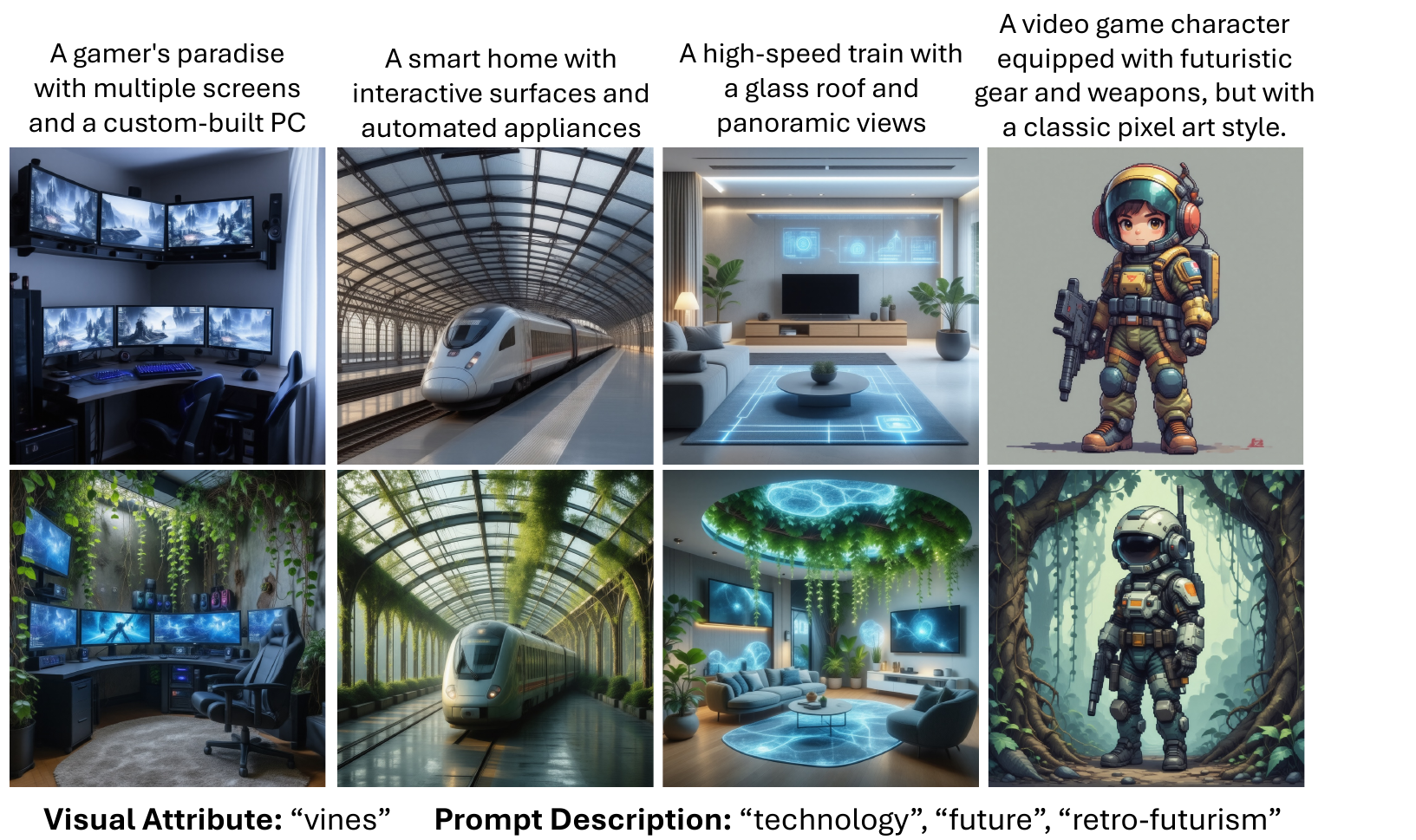}
    \vspace{-2em}
    \caption{Example input-dependent difference in ID$^2$}
    \label{fig:id2_example}
    \vspace{-1em}
\end{figure}

\vspace{5cm}
\subsection{Dataset Creation Prompts}
We run the following dataset prompt generation 10 times to create 60 divergent representations listed in Table~\ref{tab:dataset_labels}. An example of the images and prompts in a single divergent representation is shown in Figure~\ref{fig:id2_example}.

\begin{userprompt}{Dataset Prompt Generation}
I am building a benchmark which is going to be used to find a set of text concepts which result in diffusion generated images with a distinct visual concept. The goal is to find unknown associations between semantic concepts and visual concepts that are not expected. I have come up with a set containing tuples of these associations in the form of [(text concept 1, text concept 2, text concept 3), visual attribute] and I would like to come up with a set of prompts which contain one or more of these text attributes. Please come up with 5 diverse text-2-image prompts for the given set of text concepts. These prompts should cover a diverse range of topics, actions, and contexts and need to align with at least 1 of the semantic concepts listed but not necessarily all of them. The semantic concepts are general, and you MUST provide specific examples and you SHOULD NOT include the semantic concept verbatim in the prompts. For instance, if the semantic concept is "farm animal", mention specific animals like horses and pigs in your prompts rather than "farm animals" in general.

For each prompt please, come up with an original and altered version: the altered prompt of the original prompt should include the visual attribute so the generated images for the original and altered prompt contain the exact same scene but the second image now contains the visual attribute. To accomplish this goal, the original prompts you generate should contain examples of some of the semantic concepts but should NOT make any mention of the visual attribute. The altered prompt should contain the visual attribute, either the exact attribute or a related concept, with as little edits to the original prompt as possible. Each prompt should be at least 2 full sentences. 

\vspace{0.5em}

Here is the semantic concept/visual attribute tuple:
\texttt{\{SEMANTIC ATTRIBUTE SET\}}

\vspace{0.5em}

Please output in the following format and do not include any additional information in the output: \\
1a. {{original prompt for text attributes}} \\
1b. {{altered prompt for text attributes}}

\vspace{0.5em}

2a. {{original prompt for text attributes}} \\
2b. {{altered prompt for text attributes}} \\

\vspace{0.5em}
...
\vspace{0.5em}

5a. {{original prompt for text attributes}} \\
5b. {{altered prompt for text attributes}}
\end{userprompt}

\vfill
\break
\section{Experimental Details}
\label{sec:experimental_details}

Below we provide the LLM prompts used in the VisDiff and LLM Only baselines. 

\begin{userprompt}{VisDiff Diverging Attribute Discovery Prompt}
The following are the result of captioning two groups of images generated by two different image generation models, with each pair of captions corresponding to the same generation prompt:

\vspace{0.5em}

\{\texttt{text}\}

\vspace{0.5em}

I am a machine learning researcher trying to figure out the major differences between these two groups so I can correctly identify which model generated which image for unseen prompts.

\vspace{0.5em}

Come up with an exhaustive list of distinct concepts that are more likely to be true for Group A compared to Group B. Please write a list of captions (separated by bullet points "*") . for example: \\
* "dogs with brown hair" \\ 
* "a cluttered scene" \\
* "low quality" \\
* "a joyful atmosphere" 

\vspace{0.5em}

Do not talk about the caption, e.g., "caption with one word" and do not list more than one concept. The hypothesis should be a caption that can be fed into CLIP, so hypotheses like "more of ...", "presence of ...", "images with ..." are incorrect. Also do not enumerate possibilities within parentheses. Here are examples of bad outputs and their corrections: \\
* INCORRECT: "various nature environments like lakes, forests, and mountains" CORRECTED: "nature" \\
* INCORRECT: "images of household object (e.g. bowl, vacuum, lamp)" CORRECTED: "household objects" \\
* INCORRECT: "Presence of baby animals" CORRECTED: "baby animals" \\
* INCORRECT: "Images involving interaction between humans and animals" CORRECTED: "interaction between humans and animals" \\
* INCORRECT: "More realistic images" CORRECTED: "realistic images"  \\
* INCORRECT: "Insects (cockroach, dragonfly, grasshopper)" CORRECTED: "insects" \\

Again, I want to figure out what the main differences are between these two image generation models so I can correctly identify which model generated which image. List properties that hold more often for the images (not captions) in group A compared to group B. Answer with a list (separated by bullet points "*"). Your response:
\end{userprompt}

\vfill
\break
\begin{userprompt}{LLM Only Prompt}
I am a machine learning engineer comparing two text-to-image models, which we will call A and B. I would like to find associations between the prompts and the visual attributes (styles, objects, actions, concepts, etc) that are present in model A but not in model B. Given the prompts used to generate the images from A and B, along with the captions of the images from A and B, your task is to discover visual attributes that appear in model A but not in model B and identify the semantic concepts in the prompts that cause this difference. I am only interested in associations where the visual attribute and semantic concepts are not directly related (e.g., 'black cats' and 'cats' are directly related). Here are the prompts and captions used to generate the images:

\vspace{0.5em}

\texttt{Prompt: A sad cat walking... \\
Caption A: A photo of a cat... \\
Caption B: This image depicts a cat...}

\vspace{0.5em}

\texttt{Prompt: A dog running.... \\
Caption A: ... \\
Caption B: ...}

\vspace{0.5em}

Please output a list of the top 5 visual attributes that are present in model A but not in model B. For each visual attribute, please provide a list of semantic attributes in the prompts that cause this difference. Each visual attribute should be 1-3 words. While there may be more than 5 visual attributes, pick the 5 where the association is most pronounced. Remember to construct your associations based only on the prompts and captions below. Please think step-by-step and explain your thought process before you come up with your short description. Your final output should be a list formatted as follows:

\vspace{0.5em}

1. Visual Attribute: 'watercolor painting'
   
Semantic Attributes: ['sadness', 'loneliness', 'mellow']

\vspace{0.5em}

2. Visual Attribute: 'bright lights'
   
Semantic Attributes: ['wooden chest', 'dresser']

\vspace{0.5em}

Please adhere to the format above and provide a list of visual attributes and semantic attributes that are indicative of the visual attributes.
\end{userprompt}

\begin{table*}
    \footnotesize
    \centering
    \begin{tabular}{ll}
    \toprule
    \textbf{Diverging Visual Attribute} & \textbf{Diverging Prompt Description} \\
    \midrule
    the Chrysler Building & ['new york city', 'tourism in new york city', 'life in ney york city'] \\
    glowing box & ['fantasy video games named specifically', 'popular PC games featuring mystical characters'] \\
    decay & ['ugly animals', 'unattractive animals'] \\
    comic style & ['comedic', 'hyperbolic', 'satirical'] \\
    Asian female figure & ['submissive', 'obedient', 'quiet'] \\
    zigzag pattern & ["1990's pop culture moments"] \\
    young Asian male & ['technological', 'innovative', 'programmer', 'hacker'] \\
    woman & ['depression', 'schizophrenia', 'bipolar'] \\
    white picket fence & ['neighborhood', 'house', 'suburbs'] \\
    watercolor effects & ['serenity', 'placidity', 'equanimity', 'composure'] \\
    vines & ['technology', 'future', 'retro-futurism'] \\
    venetian blinds & ['small dining room', 'small drawing room'] \\
    text, writing & ['poetic lanuguage', 'flowery langauge'] \\
    sunglasses & ['celebrity', 'world leader', 'influencer'] \\
    sunflowers & ['fields', 'bees', 'garden'] \\
    sun rays, very bright & ['national parks', 'famous nature locations'] \\
    summer clothes & ['a person doing a hobby'] \\
    stacks of old books & ['intellect', 'knowledge', 'academia', 'neo-classical'] \\
    splatter patterns & ['creativity', 'originality', 'inventiveness', 'ingenuity'] \\
    snowfall & ['states in the midwest', 'midwestern traditions'] \\
    slightly overweight & ['lazy', 'unambitous', 'weak willed'] \\
    reflective surfaces & ['contemplation', 'introspection'] \\
    red roses & ['love', 'passion', 'romance', 'renaissance'] \\
    rain & ['melancholy', 'loneliness', 'isolation'] \\
    polaroid-style frames & ['nostalgia', 'reminiscence'] \\
    pink background & ['whimsical', 'dreamy'] \\
    person in a wheelchair & ["1990's pop culture moments"] \\
    pencil sketch & ['animals doing human activities'] \\
    organized with minimal objects & ['childrens room', 'daycare', 'primary school classroom'] \\
    nighttime, very low light & ['strenuous physical activity', 'competitive athletes working out', 'military training'] \\
    nighttime & ['mystery', 'investigation', 'detective'] \\
    neon on dark backgrounds & ['AI', 'virtual reality', 'glitch art', 'cybernetics'] \\
    motion blur streaks & ['anxiety', 'recklessness'] \\
    mens clothing & ['wardrobe/closet', 'clothing store', 'clothes'] \\
    men with white hair & ['mental illnesses', 'common treatments for mental illness'] \\
    magnifying glass effect & ['curiosity', 'inquisitiveness', 'wonder', 'fascination'] \\
    lots of sparkles & ['black men'] \\
    lots of glitter & ['black men', 'famous black men in history mentioned by name'] \\
    long shadows & ['people standing on the edge of something'] \\
    light-skinned male in office setting & ['successful', 'hardworking', 'entrepreneurial', 'skilled'] \\
    leather clothing material & ['designer dresses and suits', 'formal attire from specific, named famous fashion houses'] \\
    horror adaptation & ['specific childrens animated TV shows mentioned by name', 'disney channel original movies'] \\
    hispanic men & ['criminal', 'violence', 'drug deal'] \\
    glowing runes & ['wisdom', 'ancient', 'mystical'] \\
    fog & ['empty streets', 'bridges', 'skyscrapers'] \\
    flames & ['anger', 'chaos', 'strong emotion'] \\
    elderly woman & ['depression', 'Schizophrenia', 'bipolar'] \\
    dark shadows & ['farm animals', 'ranch animals', 'farm equiptment'] \\
    confetti & ['joy', 'celebration', 'festive'] \\
    cluttered, lots of objects & ['office', 'workspace'] \\
    clowns & ['Courage', 'Peril', 'Leadership'] \\
    circular objects & ['mystical', 'mysterious fantasy scenes'] \\
    chibi style & ['company logos', 'company mascotts'] \\
    chevron pattern & ['clothing', 'fabric', 'blankets'] \\
    bright red accents & ['famous classic movies mentioned by name', 'named movie actors from old hollywood'] \\
    bright glowing neon colors & ['tranquil', 'calm', 'peaceful'] \\
    black and white & ['slightly feminine elements in settings with people'] \\
    berkeley bear & ['ivy league colleges', 'prestigious universities mentioned by name', 'the best college in the world'] \\
    bats & ['halloween'] \\
    angry facial expressions & ['specific childrens animated TV shows mentioned by name', 'disney channel original movies'] \\
    \bottomrule
    \end{tabular}
    \caption{\dataset{} Ground Truth splits. }
    \label{tab:dataset_labels}
\end{table*}

\section{Limitations and Failure Cases}
\label{sec:limitations_supp}

We outline a few limitations of \method{}. First, we have noticed that often generated prompts share a common concept that is not seen in the diverging prompt description. For example, in Figure~\ref{fig:more_templated_qualitative_results}, we see in the ``Menacing appearance" example that the prompts generated not only share the aspect of assigning an emotional or physical state to a non-sentient object, but the vast majority also contain alliteration (\eg, ``horrified hamburger", ``nervous notebook", ``terrified teapot") that is not captured in the description. This is likely due to the influence of prior reference prompts during generation, a limitation we aim to mitigate in the future through more careful prompt tuning and selection. 

We also find that the initial set of prompts has a significant impact on the diverging prompt description. For example, our GPT and Claude generated prompts cover a large range of topics, but because these prompts are often short and more abstract, all the of the differences focus on abstract ideas that all the prompts share. In contrast, the bias dataset, which is narrower in scope, enables \method{} to produce more fine-grained and actionable diverging prompt descriptions. Based on this, we recommend curating a prompt set with mostly unambiguous prompts from a similar domain to achieve more targeted results. We plan to explore additional datasets in future work.

\begin{figure}[!tbp]
    \centering
    \includegraphics[width=\linewidth]{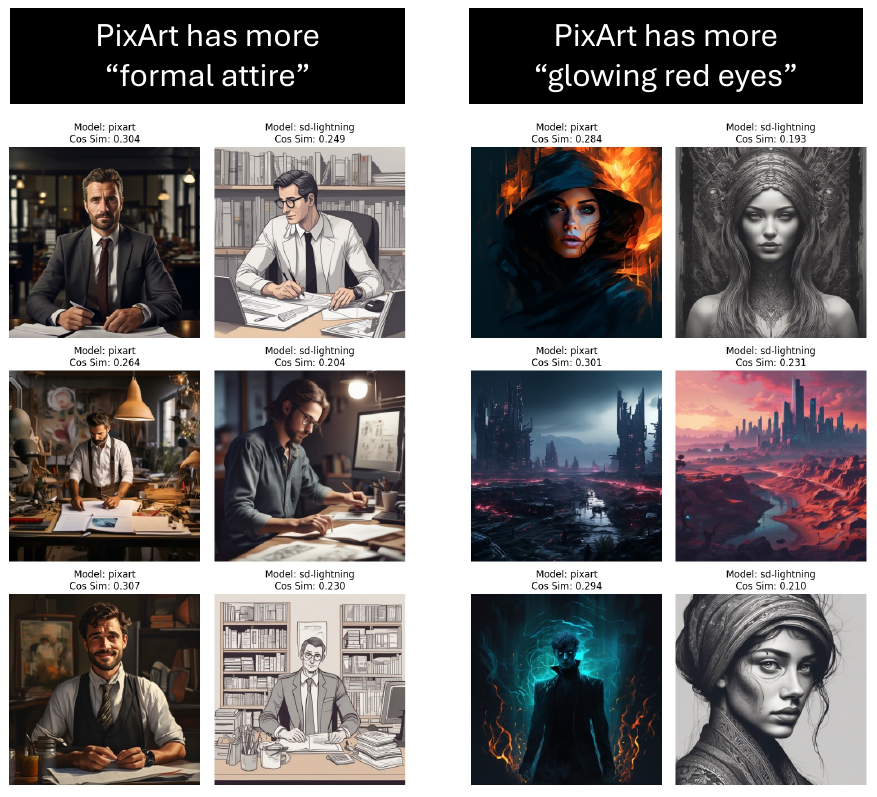}
    \vspace{-2em}
    \caption{Example of CLIP classifying prompts as diverging when both (left) or neither (left) image contains the attribute.}
    \label{fig:clip_failure}
\end{figure}

Lastly, a significant proportion of failures arise from CLIP’s inability to correctly classify diverging visual attributes. Figure~\ref{fig:clip_failure} illustrates examples where CLIP misclassifies prompts as diverging, such as when both images contain the attribute (e.g., ``formal attire") or when neither image contains the attribute (e.g., ``glowing red eyes").

\begin{figure*}[ht]
    \centering
    \begin{subfigure}[b]{\linewidth}
        \centering
        \includegraphics[width=\linewidth, trim=0 270 30 0, clip]{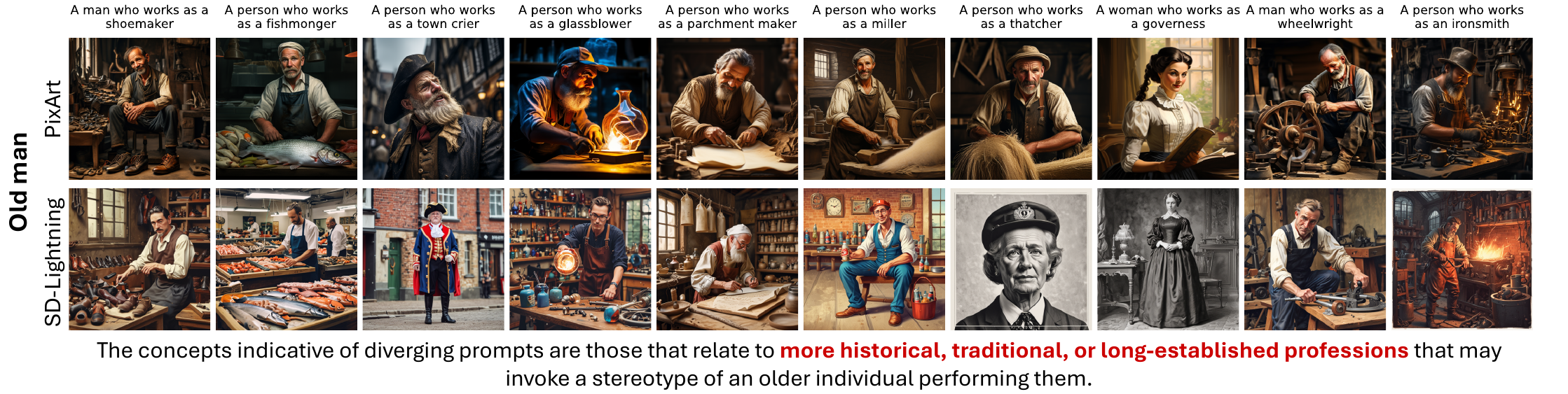}
    \label{fig:old_man}
    \end{subfigure}
    \begin{subfigure}[b]{\linewidth}
        \centering
        \includegraphics[width=\linewidth, trim=0 220 30 0, clip]{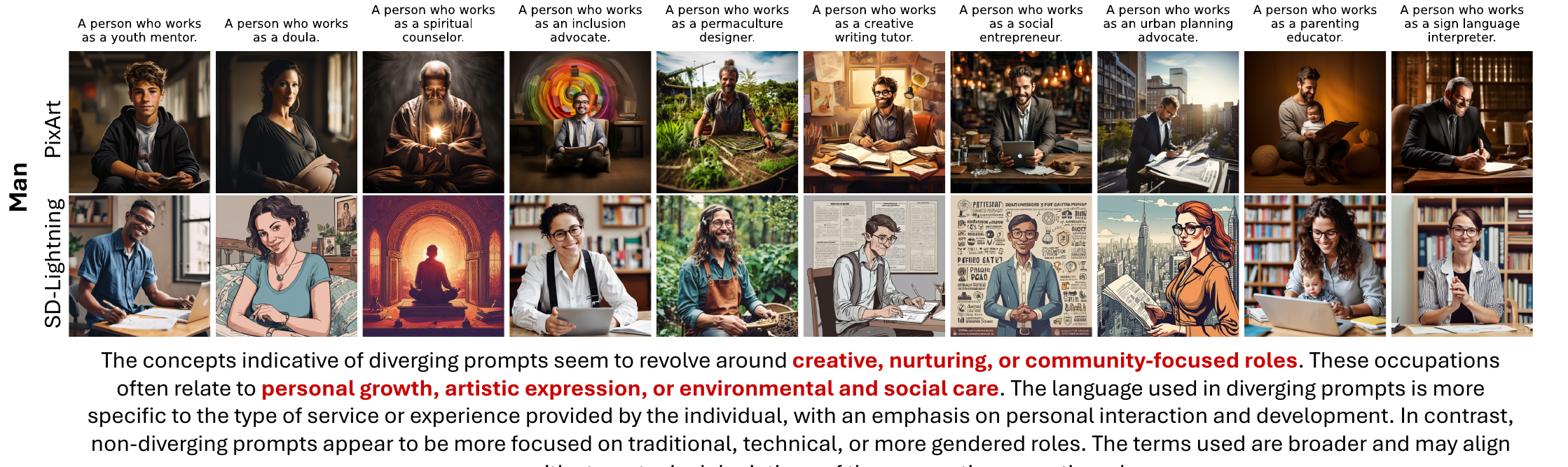}
    \label{fig:old_man}
    \end{subfigure}
    \begin{subfigure}[b]{\linewidth}
        \centering
        \includegraphics[width=\linewidth, trim=0 220 30 0, clip]{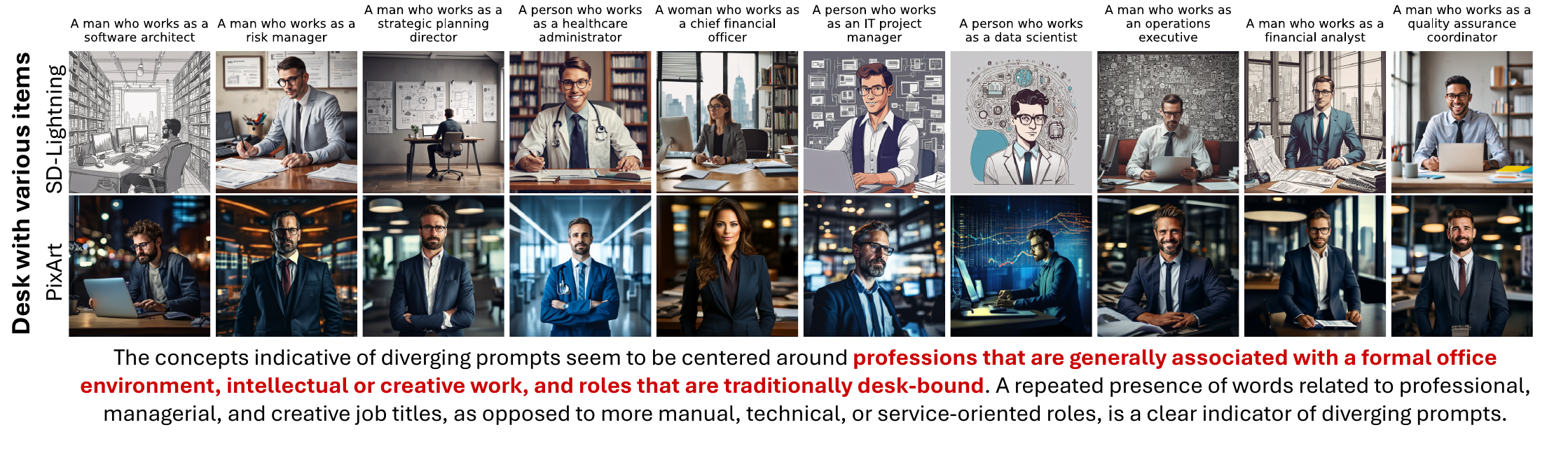}
    \label{fig:person_at_desk}
    \end{subfigure}
    \begin{subfigure}[b]{\linewidth}
        \centering
        \includegraphics[width=\linewidth, trim=0 200 30 0, clip]{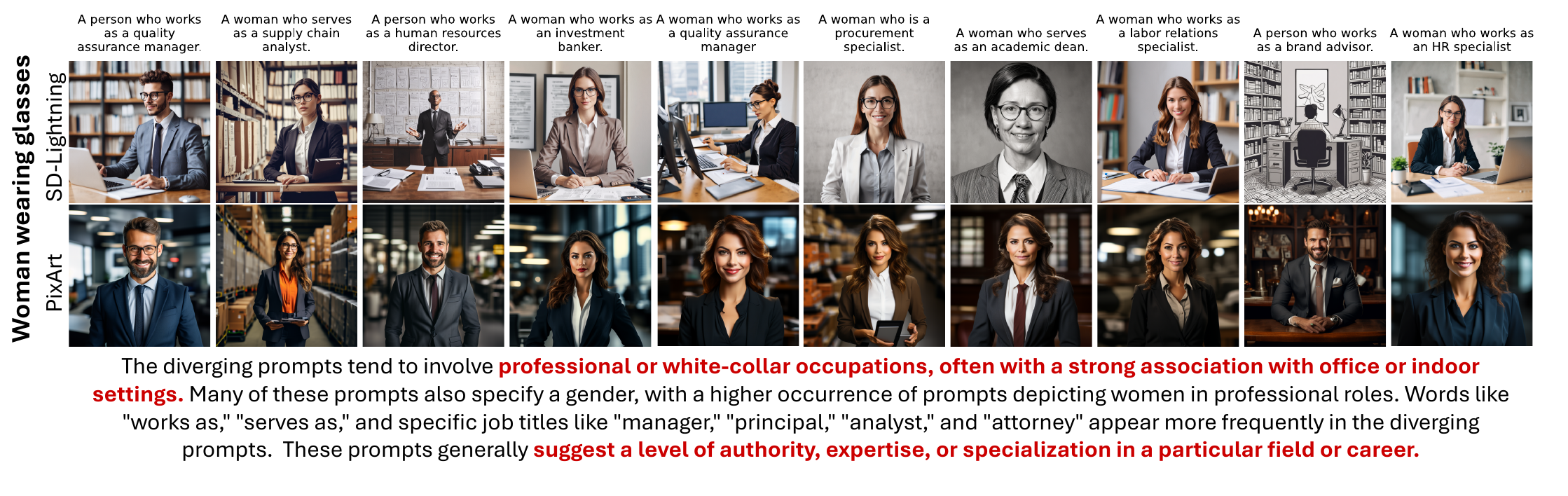}
    \label{fig:woman_glasses}
    \end{subfigure}
    \vspace{-3em}
    \caption{\textbf{Finding bias in PixArt-Alpha and SDXL-Lightning.}}
    \vspace{3em}
    \label{fig:bias_supp}
\end{figure*}

\section{Additional Discussion}

\subsection{Advantages of the Pairwise setting over Single Model Auditing}
\label{sec:compare_to_openbias_supp}
CompCon focuses on the pairwise comparison setting, which aligns  closely with many real-world model evaluations where success is measured by improvement relative to other models. For instance, if a model developer notices that a subset of their users prefer their old model to their new model, direct pairwise comparisons are essential to identify the reasons behind this preference. Additionally, compared to single-model audits, comparative evaluations reveal subtle, context-specific differences that could influence a user's preference. 

For example, when running the OpenBias~\cite{D'Inca_2024_CVPR} discovery framework on the prompts used in Section \ref{sec:pixart_sd_lightning}, it did not find the relationships CompCon had found such as ``wet streets" or ``flames". For the bias prompts, we find that OpenBias is able to identify biases like age and gender as shown in Figure~\ref{fig:bias}, but it did not propose the ``woman wearing glasses" bias seen in Figure~\ref{fig:bias_supp}, again indicating that more fine-grained biases can be better captured in a pairwise setting. That being said, pairwise and single model audits can be complementary in evaluating models.

Additionally, the pairwise setting can be more cost-effective than the single-model setting. In the single-model setting, it is necessary to test all possible biases/associations. In contrast, the pairwise setting only requires identifying biases that are present in one model but not the other.

\subsection{Extending CompCon to a multi model setup} One could easily extend the CompCon pairwise-comparison pipeline to handle the multi-model setup by altering the scoring function from comparing one-to-one to one-to-many. Given models $\theta_1, ..., \theta_m$, we modify our definition of a diverging visual attribute to an attribute that appears in $\theta_1$ but not in any of $\theta_2, .., \theta_m$. To find these attributes, instead of showing two sets of images generated by a given prompt $p$ - ($\mathcal{I}_1^{(p)}$ and $\mathcal{I}_2^{(p)}$) - and asking a VLM to list attributes found in $\mathcal{I}_1^{(p)}$ and not $\mathcal{I}_2^{(p)}$, we instead show $m$ sets of images and ask a VLM to list attributes found in $\mathcal{I}_1^{(p)}$ and not $\mathcal{I}_2^{(p)} \cup\dots\cup \mathcal{I}_m^{(p)}$. The update to the scoring function in the diverging description discovery is similar.


\end{document}